\documentclass[lettersize,journal,x11names]{IEEEtran}
\usepackage{amsmath,amsfonts}
\usepackage{algorithmic}
\usepackage{algorithm}
\usepackage{array}
\usepackage[caption=false,font=normalsize,labelfont=sf,textfont=sf]{subfig}
\usepackage{textcomp}
\usepackage{stfloats}
\usepackage{url}
\usepackage{verbatim}
\usepackage{graphicx}
\usepackage{color}
\usepackage{xcolor}
\usepackage{cite}
\usepackage{color}
\usepackage{amssymb}
\usepackage{amsmath}
\usepackage{multirow}
\usepackage{makecell} 
\usepackage{gensymb}
\usepackage{hyperref}
\begin{document}

\title{A Non-autoregressive Multi-Horizon Flight Trajectory Prediction Framework with Gray Code Representation}

\author{Dongyue Guo, Zheng Zhang, Zhen Yan, Jianwei Zhang, Hongyu Yang and Yi Lin
        % <-this % stops a space
\thanks{This work was supported by the National Natural Science Foundation of China (NSFC) under grants No. 62371323, U2333209, and U20A20161. (Corresponding author: Yi Lin; e-mail: yilin@scu.edu.cn)}

\thanks{Dongyue Guo, Zheng Zhang, Jianwei Zhang, Hongyu Yang and Yi Lin are with the College of Computer Science, Sichuan University, Chengdu 610000, China.} 
\thanks{Zhen Yan is with the State Key Laboratory of Air Traffic Management System, Beijing 100000.}}

% \markboth{IEEE Transactions on Intelligent Transportation Systems}%
% \markboth{The Preprint Version}
% {Guo \MakeLowercase{\textit{et al.}}: A Non-autoregressive Multi-Horizon Flight Trajectory Prediction Framework using Gray Code Representation}


\maketitle

\begin{abstract}
Flight Trajectory Prediction (FTP) is an essential task in Air Traffic Control (ATC), which can assist air traffic controllers in managing airspace more safely and efficiently. 
Existing approaches generally perform multi-horizon FTP tasks in an autoregressive manner, thereby suffering from error accumulation and low-efficiency problems. 
In this paper, a novel framework, called FlightBERT++, is proposed to i) forecast multi-horizon flight trajectories directly in a non-autoregressive way, and ii) improve the limitation of the binary encoding (BE) representation in the FlightBERT framework. 
Specifically, the proposed framework is implemented by a generalized encoder-decoder architecture, 
in which the encoder learns the temporal-spatial patterns from historical observations and the decoder predicts the flight status for the future horizons. 
Compared to conventional architecture, an innovative horizon-aware contexts generator is dedicatedly designed to consider the prior horizon information, which further enables non-autoregressive multi-horizon prediction. 
Additionally, the Gray code representation and the differential prediction paradigm are designed to cope with the high-bit misclassifications of the BE representation, which significantly reduces the outliers in the predictions. 
Moreover, a differential prompted decoder is proposed to enhance the capability of the differential predictions by leveraging the stationarity of the differential sequence. 
Extensive experiments are conducted to validate the proposed framework on a real-world flight trajectory dataset. 
The experimental results demonstrated that the proposed framework outperformed the competitive baselines in both FTP performance and computational efficiency. 
The code is publicly available at: https://github.com/gdy-scu/FlightBERT\_PP\_V2. 
\end{abstract}


\section{Introduction} \label{sec1}
\IEEEPARstart{F}{light} Trajectory Prediction (FTP) is essential for Air Traffic Management (ATM), 
enabling many critical applications to help Air Traffic Controllers (ATCOs) manage airspace more safely and efficiently, 
such as traffic flow prediction \cite{LIN2019105113,YAN2023101924}, conflict detection \cite{53645, ChenG020, 8561168}, and arrival time estimation \cite{9670444, wang2018hybrid}. 
In particular, the FTP is one of the fundamental techniques to support the Trajectory-Based Operation (TBO) that is being promoted by both the Single European Sky ATM Research (SESAR) and the New Generation Air Transportation Systems (NextGen) \cite{faa2010common}. 
In this context, the FTP technique has gathered more attention from researchers and achieved significant progress in the past decade.  

In general, FTP in the Air Traffic Control (ATC) domain aims to forecast the flight status in future time steps according to the observed historical flight trajectories. 
Accurate FTP requires high prediction capacity to capture the temporal-spatial dependencies between the observations and predictions. 
To investigate the robust representations for the trajectory attributes, the binary encoding (BE) is proposed to encode the attributes of the trajectory points into a set of binary vectors in our previous FlightBERT framework \cite{9945661}, further formulating the TP task as multiple binary classification (MBC) problem. 
Benefiting from the innovative idea and dedicatedly network design, the FlightBERT framework not only enhances the semantic representation of the trajectory points, but also avoids the vulnerability impacted by the normalization algorithms. 
However, two primary limitations should be further addressed to enhance the performance of the FTP task. 

\begin{itemize}
    \item A limitation of the Binary Encoding (BE) representation is that the high-bit misclassification will lead to outliers in the predictions \cite{9945661}. 
    For instance, given the BE representation "0110 0100" (decimal 100), the absolute error is 128 (decimal) if the prediction error occurs in the $8^{th}$ bit ("1110 0100") 
    while that is 1 (decimal) if the prediction error occurs in $1^{st}$ bit ("0110 0101"). 

    \item The FlightBERT performs multi-horizon prediction recursively, i.e., predicts the flight status of the next time step based on observation and 
    iteratively applies the predicted values as pseudo-observation to obtain multi-horizon predictions. 
    On the one hand, it is easy to suffer from larger accumulative errors since the prediction errors in the pseudo-observation will be accumulated during the recursive inference process. 
    On the other hand, the computational efficiency is limited by the prediction horizon due to the step-by-step prediction paradigm. 
    As the prediction horizon increases, the inference speed will experience a severe reduction, leading to unacceptable delays for real-time applications (e.g., conflict detection).
\end{itemize}

Focusing on the aforementioned challenges, in this paper, a non-autoregressive multi-horizon flight trajectory prediction framework, named FlightBERT++, is proposed to improve the performance of the FTP task. 
Thanks to the superior trajectory representation ability of the BE, the proposed FlightBERT++ framework inherits the powerful representation capacity from the FlightBERT, and is also implemented based on the MBC paradigm. 

In order to overcome the outlier predictions resulting from the high-bit misclassification in the BE representation, in this work, an innovative Gray Code (GC) representation, and a differential prediction paradigm are incorporated into the FlightBERT++ framework. 
Specifically, i) the Gray code is employed to alternate the BE representation to ensure steady bit transition patterns of the trajectory attributes between the adjacent time steps. 
ii) instead of predicting the original values, the differential values of the trajectory attributes are formulated as the output objective in the proposed framework, which can be encoded into GC representations in fewer bits. 
In this way, the FlightBERT++ framework is able to effectively mitigate the occurrence of extremely unreliable outliers in the predictions.

To achieve high-accuracy and -efficiency multi-horizon FTP prediction, a novel sequence-to-sequence (Seq2Seq) based encoder-decoder architecture is proposed to implement the FlightBERT++ framework.
Specifically, the proposed framework is constructed by cascading the trajectory encoder, Horizon-aware Context Generator (HACG), and Differential-prompted Decoder (DPD).  
Firstly, the trajectory encoder encodes the observations into a trajectory-level representation, while the HACG is designed to produce multi-horizon context representations of the predicted horizons. 
Benefiting from the proposed HACG module, unlike conventional Seq2Seq architecture, the proposed framework can generate multi-horizon predictions directly (non-autoregressive) rather than perform recursive inference. 
In succession, the differential-prompted decoder is innovatively proposed to perform the high-confidence predictions based on the multi-horizon context representations, which employ the differential sequence of the observations as the prompting. 
In the proposed framework, the Transformer blocks are applied to build the backbone network for both the trajectory encoder and DPD, which allows the network to perform non-autoregressive inference in the temporal modeling process. 
In this way, the proposed framework can not only mitigate the error accumulation but also yield a substantial improvement in computational efficiency for multi-horizon FTP tasks. 

This paper is an extended version of our previous work published at the AAAI Conference \cite{guo2023flightbert++}, which improves our original work as follows:
i) An innovative Gray Code representation is proposed to encode the trajectory attributes to tackle the high-bit misclassification problems, which achieves 29.0\% and 21.8\% mean deviation error reduction in 1 and 15 horizon predictions. 
ii) Two more competitive baselines are considered to validate the proposed approach, including the latest work WTFTP \cite{Zhang_2023} and the typical multi-horizon forecast architecture Transformer-Seq2Seq \cite{VaswaniSPUJGKP17}. 
iii) Additional technical details, neural architecture and related works are provided to enhance the systematicity and reproducibility of this work.
iv) More insightful experiments (as well as ablation studies), results, visualization and analyses for different flight phases are also provided in this version.

The proposed framework is evaluated on a real-world flight trajectory dataset from an industrial ATC system.  
To validate the effectiveness and efficiency of the proposed framework, a total of 7 competitive baselines are selected to conduct comprehensive comparisons.  
In addition, extensive ablation studies and insightful analysis are also performed to confirm all proposed technical improvements. 
The experimental results consistently demonstrated that the proposed framework efficiently addresses the outliers and error accumulation problems, outperforming the baselines in both FTP accuracy and efficiency. 
In summary, the contributions and novelty of this work are listed as follows: 

\begin{itemize}
    \item A flight trajectory prediction framework, called FlightBERT++, is innovatively proposed to perform high-accuracy and -efficiency multi-horizon forecasting in a non-autoregressive manner. 
    \item The Gray code representation and differential prediction paradigm is dedicatedly designed to overcome the limitations of the high-bit misclassification in BE representation. 
    \item A HACG is innovatively proposed to generate multi-horizon context representations by leveraging prior horizon knowledge, which is the key idea to supporting non-autoregressive predictions. 
    \item Considering the stationarity of the differential sequence in flight trajectory, a differential-prompted decoder is proposed to assist in learning the transition patterns of the trajectory sequence, which further improves the performance of the FlightBERT++.  
    \item We conducted extensive comparison experiments on the real-world trajectory dataset to demonstrate the effectiveness of the proposed frameworks. 
    In addition, some diagnostic studies and analyses are applied to explore the robustness and superiority of the proposed framework. 
\end{itemize}

The rest of this paper is organized as follows. 
The related literature and works of the FTP task are briefly reviewed in Section \ref{sec2}. 
Section \ref{sec3} detailed presents the methodologies of the proposed FlightBERT++ framework. 
The dataset, experimental settings, baselines, and evaluation metrics are described in Section \ref{sec4}. 
The experimental results, visualization, and discussions are presented in Section \ref{sec5}. 
Finally, the work of this paper is concluded in Section \ref{sec6}.

\section{Related Work} \label{sec2}
\subsection{Flight Trajectory Prediction} \label{sec2.1}
Currently, the short FTP tasks are commonly formulated as a time series forecasting (TSF) problem, which predicts future flight trajectories based on historical observations. 
According to modeling approaches, the FTP methods can be classified into three distinct categories: physical models, filter-based models, and data-driven models. 

\subsubsection{Physical models}
Physical models typically establish a set of mathematical equations based on kinematics and aerodynamics assumptions to estimate the future flight status \cite{9780784, 2014-2198, 9009, faa2010common}. 
In this paradigm, the aircraft is generally modeled as a point mass and many factors are considered and incorporated into the physical models to perform more accurate predictions,
such as aircraft dynamics, aircraft performance, and atmospheric conditions \cite{2006-6098, porretta2008, 4391830}. 
Although modern physical models can obtain the Aircraft Performance Models (APMs) and Operating Procedure Coefficients (OPCs) from the public database, such as Base of Aircraft Data (BADA) \cite{nuic2010user},  
certain sensitive information is still not disclosed to researchers publicly. 
In \cite{sun2016modeling}, an aircraft takeoff mass inferring method was proposed to enhance the APMs by studying the kinetic model at the lift-off moment and observing the aircraft motion on the runway.  
To improve the trajectory prediction accuracy, an unknown point-mass model parameters learning approach was proposed to estimate the mass and thrust of the aircraft \cite{ALLIGIER201345}. 
These unknown parameters were learned from historical observations and adjusted by fitting the modeled specific power to the observed energy rate. 

It is evident that physical models have been extensively investigated over the past few decades and derived various applications in modern ATC and avionics systems. 
However, the performance of physical models is highly sensitive to the specific performance parameters of the aircraft and typically requires modeling tailored to different aircraft types. 
In addition, limited by the fixed inference rules, physical models usually suffer from performance degradation during flight maneuvering phases due to more complex transition patterns. 

\subsubsection{Filter-based models} 
Filter-based models are the classical FTP approaches that estimate the flight trajectory iteratively via a predefined system model by considering real-time measurements. 
These approaches are generally implemented based on the Kalman Filter (KF) \cite{1960A} or its variants, which continuously corrects the trajectory estimates based on the observed measurements, 
providing a more accurate and dynamically updated prediction \cite{6972562, 58508}. 
In practice, filter-based models are commonly used for short-term FTP tasks or target tracking tasks to mitigate errors or uncertainties in the measurements. 
To improve the FTP performance in conflict detection and resolution applications, the KF algorithm was employed to provide the aircraft state estimation in \cite{xi2008simulation}. 
The estimated flight state was further used in flight intent determination and predicted the future trajectory. 
In \cite{yan2013application}, the Unscented Kalman Filter (UKF) algorithm was applied to simulate and predict target trajectory and achieve the flying target tracking. 
Towards the challenges of target motion-mode uncertainty and nonlinearity, the Multiple Model (MM) methods have been generally considered the mainstream approach in the field of target tracking \cite{1561886, 8569251}. 
Moreover, the hybrid estimation algorithm was also developed to track aircraft in terminal airspace \cite{40127}. 

Compared to physical models, since the dynamically updated nature of parameters in the system model, Filter-based models can adapt to various flight scenarios flexibly. 
However, these approaches are not suitable for multi-step prediction tasks as they rely on real-time observations to update the parameters of the system model.

\subsubsection{Data-driven models}
Data-driven models aim to learn flight transition patterns from historical data, enabling high-precision predictions. 
These approaches usually build FTP models using machine learning or deep learning techniques and fit the model parameters in the training process \cite{01652041, 9166485}. 
In \cite{2939694}, a stochastic FTP method was proposed to align trajectories into a 3D grid network and utilize the Hidden Markov Model (HMM) to account for environmental uncertainties.  
In \cite{2013-4782}, a machine learning approach was proposed to perform FTP tasks using the stepwise regression algorithm for arrival time prediction. 
Furthermore, the Gaussian Mixture Model (GMM) \cite{wiest2012probabilistic}, Gaussian Processes \cite{yan2017probabilistic}, Support Vector Machine (SVM) \cite{jiang2018svm}, and various variant algorithms were also introduced and adapted into FTP tasks.

With the development of the deep neural networks (DNNs), various sequential modeling neural architectures were employed to implement the FTP tasks, such as Recurrent Neural Networks (RNNs), Transformer \cite{VaswaniSPUJGKP17}, and Convolutional Networks (CNN).
In \cite{wu20204d}, the backpropagation (BP) neural network was employed to perform the high-precision four-dimensional TP task. 
The Long Short Term Memory (LSTM) network was also widely introduced into FTP tasks due to its significant temporal modeling capabilities \cite{ShiXPYZ18}. 
A hybrid architecture was proposed to capture the spatial-temporal features of the trajectory data by combining CNN and LSTM network \cite{9145522}. 
Moreover, the prior physics knowledge was also incorporated into deep learning based FTP approaches to enhance the model performance in different flight phases \cite{zhang2022phased, ShiXP21}.
In addition, the data mining and Bayesian neural network were also introduced to model FTP tasks in some ATC applications \cite{6246959, zhang2020bayesian, pang2021data}. 

In general, deep learning-based models have emerged as the predominant approach in modern FTP applications owing to their remarkable performance. 
Nevertheless, it is worth noting that data-driven models necessitate high-quality and representative training data to achieve desired performance.

\subsubsection{Multi-horizon Forecasting} \label{sec2.2}
Currently, most existing short-term FTP works focused on capturing the flight transition patterns and performing iteratively multi-horizon inference using predicted results in the last time-step. 
However, the iteratively multi-horizon forecasting paradigm suffered from error accumulation problems due to the pseudo-observations as inputs. 
Towards this gap, the Seq2Seq architectures were applied to develop the multi-horizon FTP tasks, which are able to generate the trajectory of multiple time steps directly. 
In \cite{9704880}, an encoder-decoder architecture implemented on CNN and Gated Recurrent Units (GRU) was proposed to conduct FTP models, which exhibit a remarkable performance in multi-horizon predictions. 
Despite the partial alleviation of error accumulation by Seq2Seq models, the performance and inference speed remain limited in multi-horizon predictions due to the autoregressive inference principles. 
Therefore, this work aims to not only improve the limitations of the BE representations but also develop a non-autoregressive framework to perform high-precision multi-horizon FTP tasks. 

\section{Methodology} \label{sec3}
\subsection{Problem Formulation} \label{sec3.1}
In general, the short-term FTP task can be formulated as a multi-variable time-series forecasting problem, i.e., predicting the trajectory point (composed of several trajectory attributes) of future horizons based on historical observations. 
Let an observed trajectory point $p_t$ in timestamp $t$, the multi-horizon FTP aims to forecast the $P_{t+1:t+n} = \{p_{t+1}, p_{t+2}, ..., p_{t+n}\}$ based on the observation sequence $O_{t-k+1:t} = \{p_{t-k+1}, ..., p_{t-1}, p_t\}$. 
Mathematically, it can be abstracted as Eq. (\ref{eq1}): 
\begin{equation} \label{eq1}
        P_{t+1:t+n} = \{p_{t+1}, p_{t+2}, ..., p_{t+n}\} =  \mathcal{F} (O_{t-k+1:t})
\end{equation}
\noindent where the $n, k$ are the number of prediction horizons and the observed sequence length, respectively. 
$\mathcal{F}(\cdot)$ denotes the learnable FTP model. 
Compared to the conventional approaches, instead of performing the prediction iteratively, $\mathcal{F}(\cdot)$ outputs the multi-horizon predictions directly in the proposed FlightBERT++ framework. 

In this work, the trajectory point $p_t$ is formulated as a collection of six key attributes that mainly depict the aircraft status. The definition is presented as follows. 
\begin{equation} \label{eq2}
        p_t = [Lon_t, Lat_t, Alt_t, {Vx}_t, {Vy}_t, {Vz}_t]
\end{equation}
\noindent where the $Lon_t, Lat_t, Alt_t, {Vx}_t, {Vy}_t, {Vz}_t$ represents the longitude, latitude, altitude, and velocity in $x, y, z$ dimensions of the trajectory point, respectively. 
The $x, y, z$ dimensions of the velocity correspond to the longitudinal, latitudinal, and altitudinal attributes, respectively.

\subsection{Gray Code Representation and Differential Prediction} \label{sec3.2-1}
In our previous work, i.e., FlightBERT \cite{9945661}, the binary encoding representation was proposed to convert the scalar trajectory attributes into high-dimensional $n \mbox{-} hot$ vectors to boost the FTP tasks. 
Consequently, FlightBERT formulated the FTP as the multi-binary classification (MBC) task. 
As mentioned in Section \ref{sec1}, although FlightBERT demonstrated great performance improvements compared to conventional FTP approaches,
the susceptibility of the BE representation to high-bit misclassifications inevitably leads to some outliers in the prediction results.

In practice, the flight transition patterns of civil aviation aircraft exhibit prominent stability in terms of the longitude and latitude measurements, and trend variations in trajectory attributes typically fluctuate in a small range. 
In the majority of instances involving the BE representation, even marginal alterations (e.g., a decimal increment of 1) in trajectory attributes in the next time step yield significant disparities between resulting $n \mbox{-} hot$ vectors, such as decimal 127 (BE: 0111 1111) and 128 (BE: 1000 0000). 
It is believed that the significant changes in bit transition patterns in the BE representations bring challenges to the model to effectively capture the trajectory dynamics.

\begin{figure}[htbp]
	\centering
	\includegraphics[width=3.5 in]{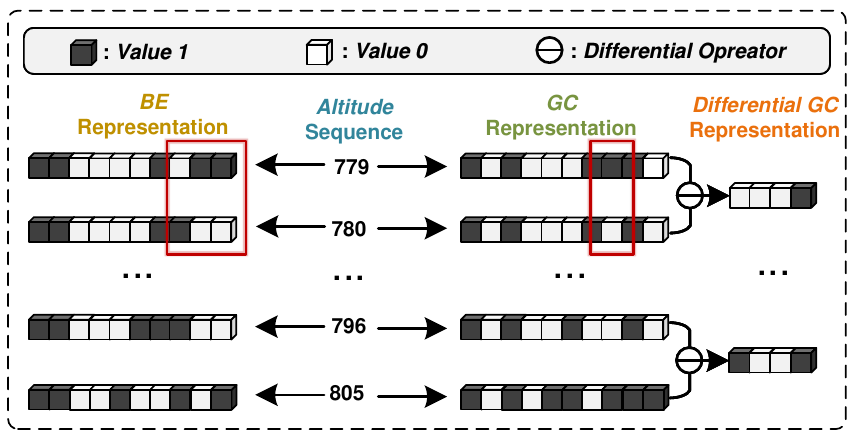}
	\caption{Comparison of the BE, GC, and Differential GC representations. The examples of bit state changes are marked by the red rectangles.} 
	\label{fig_gc}
\end{figure}

The Gray code, also known as reflected binary code, is an ordering of the binary numeral system, such that two successive values differ in only one bit (binary digit). 
Inspired by this, the Gray Code (GC) representation is proposed to alternate BE to encode the trajectory attributes into $n \mbox{-} hot$ vectors, which minimizes the bit variations between consecutive time steps.  
Specifically, let an observed altitude sequence $Alt = [780, 781, ...., 796, 805]$, the encoding process and a detailed comparison between BE and GC representation are depicted in Fig. \ref{fig_gc}.
Notably, when the values increment from 779 to 780 (decimal), the GC representations exhibit only the $3^{rd}$ bit transitioning from 1 to 0, whereas a total of 3 bits have state changes in BE representations.
Leveraging this inherent characteristic of GC, the model is expected to effectively capture the bit transition patterns, thereby alleviating high-bit misclassifications. 

Moreover, to further mitigate the high-bits misclassifications of the BE representation, the differential prediction paradigm is introduced into the FlightBERT++ framework, 
i.e., the objective of the decoder is to predict the differential GC representation instead of the GC representation raw absolute values. This design has the following merits:
i) Since the trend variations of differential in the observation sequence are with higher stationary compared to the raw absolute sequence, the dynamics of the trajectory sequence can be effectively captured by the FTP model.
ii) By using differential values, their numerical range is smaller, allowing for the encoded GC representation with fewer bits. Consequently, even if there are misclassifications in the higher bits, the error can be limited within a certain range, thereby eliminating the outlier in predictions. 
Therefore, it is believed that predicting the differential values in GC representation is an effective way to improve the performance of the FTP task. 

\begin{figure*}[htbp]
	\centering
	\includegraphics[width=7 in]{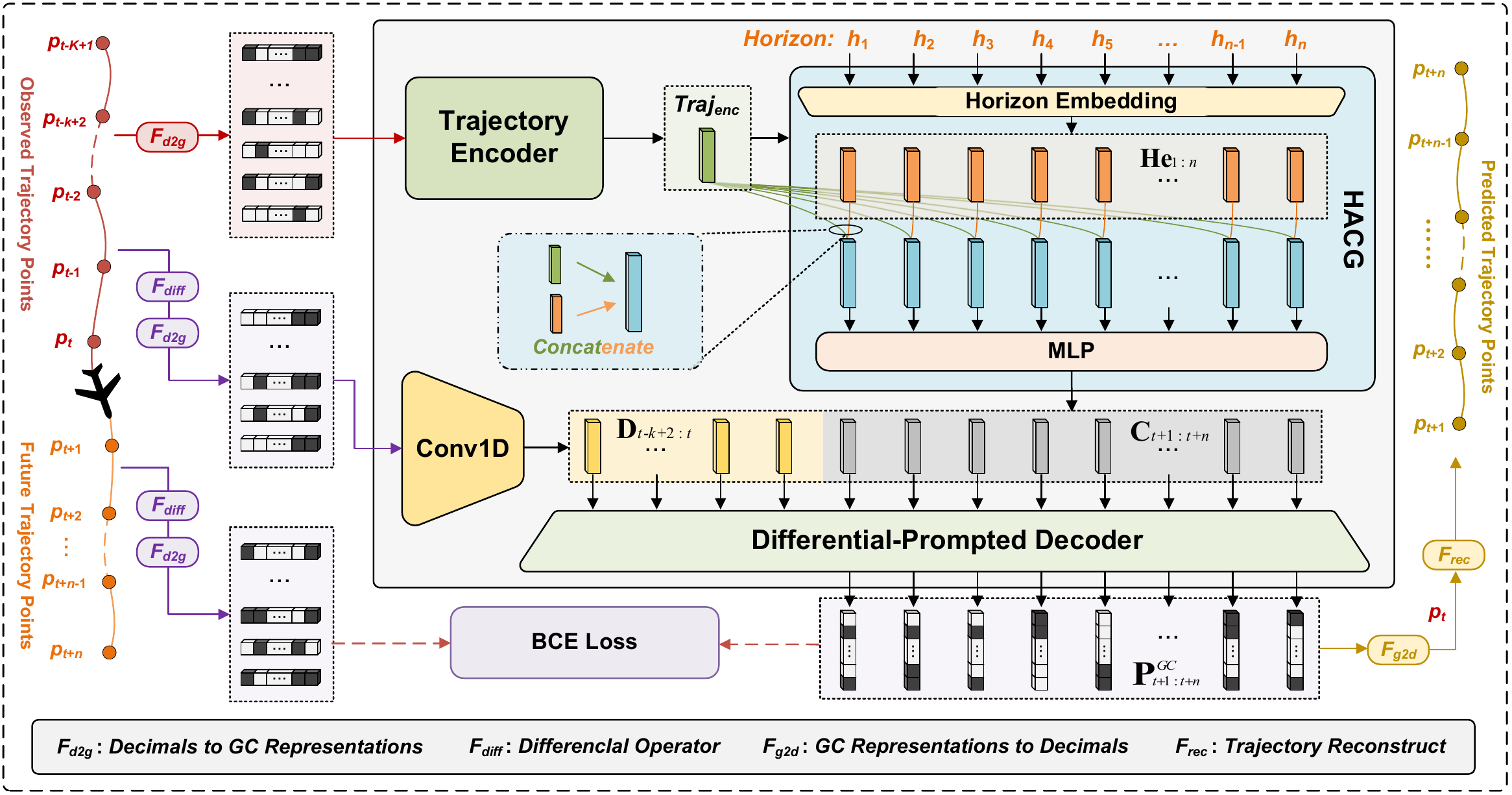}
	\caption{Overview of the proposed FlightBERT++ framework.} 
	\label{fig1}
\end{figure*}

\subsection{The Proposed Framework} \label{sec3.2}
The proposed framework is illustrated in Figure \ref{fig1}. 
The neural architecture of the proposed framework is cascaded by a trajectory encoder, horizon-aware context generator, and Differential-prompted Decoder (DPD). 
Given the observed trajectory sequence $O_{t-k+1:t}$, the object of the trajectory encoder is to learn the temporal-spatial correlations of the observations and abstract them to a high-dimensional representation $\mathbf{Traj}_{enc}$, as Eq. (\ref{eq3}). 
\begin{equation} \label{eq3}
        \mathbf{Traj}_{enc} = \operatorname{TrajectoryEncoder}(O_{t-k+1:t})
\end{equation}

In succession, the HACG is designed to generate the multi-horizon context representations $C_{t+1:t+n} = \{c_{t+1}, c_{t+2}, ..., c_{t+n}\}$ by both considering the prior prediction horizon encodings $\mathbf{H} = \{h_{1}, h_{2}, ..., h_{n},\}$ and the high-level representation $\mathbf{Traj}_{enc}$. 
\begin{equation} \label{eq4}
        \mathbf{C}_{t+1:t+n} = \operatorname{HACG}([\mathbf{Traj}_{enc}, \mathbf{H}])
\end{equation}

The DPD receives two input vectors, i.e., the differential embeddings $\mathbf{D}_{t-k+2:t}$ and multi-horizon context representations $\mathbf{C}_{t+1:t+n}$. 
Specifically, the differential embeddings $\mathbf{D}_{t-k+2:t}$ of the $O_{t-k+1:t}$ are extracted by a Conv1D neural network, which severs as the prompting to learn the transition patterns of the differential sequence. 
Then, the extracted $\mathbf{D}_{t-k+2:t}$ and multi-horizon context representations $\mathbf{C}_{t+1:t+n}$ are jointly fed into the differential-prompted decoder to generate the final outputs as Eq. (\ref{eq5}), 
where the $\mathbf{P}_{t+1: t+n}^{GC}$ is the GC representation of the predicted differential sequence of future trajectory. 
Finally, the $\mathbf{P}_{t+1: t+n}^{GC}$ is transformed into decimals to reconstruct the predicted trajectory based on the $O_t$. 

\begin{equation} \label{eq5}
        \mathbf{P}_{t+1: t+n}^{GC} = \operatorname{DPD}([\mathbf{D}_{t-k+2:t}, \mathbf{C}_{t+1:t+n}])
\end{equation}

Note that the inputs and the outputs of the proposed framework are both the GC representations of the trajectory attributes. 
Therefore, the optimizing objective of the proposed framework is also formulated as the MBC task. 
The FlightBERT++ framework is trained using the Binary Cross Entropy (BCE) loss function, as that in conventional MBC tasks. 
More details of the BE representations and MBC-based FTP task can be found in our previous work \cite{9945661}. 

Compared to existing multi-horizon FTP approaches, the proposed FlightBERT++ framework (i) innovatively designed a HACG to generate multi-horizon contexts directly, 
(ii) employ the Transformer-based architecture to conduct the backbone network of the trajectory encoder and differential-prompted decoder, which are the core ideas in enabling non-autoregressive prediction.

\subsection{Trajectory Encoder} \label{sec3.3}
As illustrated in Figure \ref{fig1}, the trajectory encoder is composed of three modules, including the Conv1D-based Trajectory Point Embedding (TPE) module, Transformer-based temporal modeling module, and Attention-based Sequence Aggregation (ASA) module. 
Specifically, the Conv1D-based TPE module projects the GC representation into high-dimensional embedding space to learn discriminative spatial features of the trajectory points, while the Transformer-based module is employed to capture the temporal correlations among the observation sequence. 
In succession, the outputs of the Transformer module are further fed into the ASA module to generate the trajectory-level embedding and extract the semantic representation over the whole observation sequences. 

\begin{figure*}[htbp]
	\centering
	\includegraphics[width=6 in]{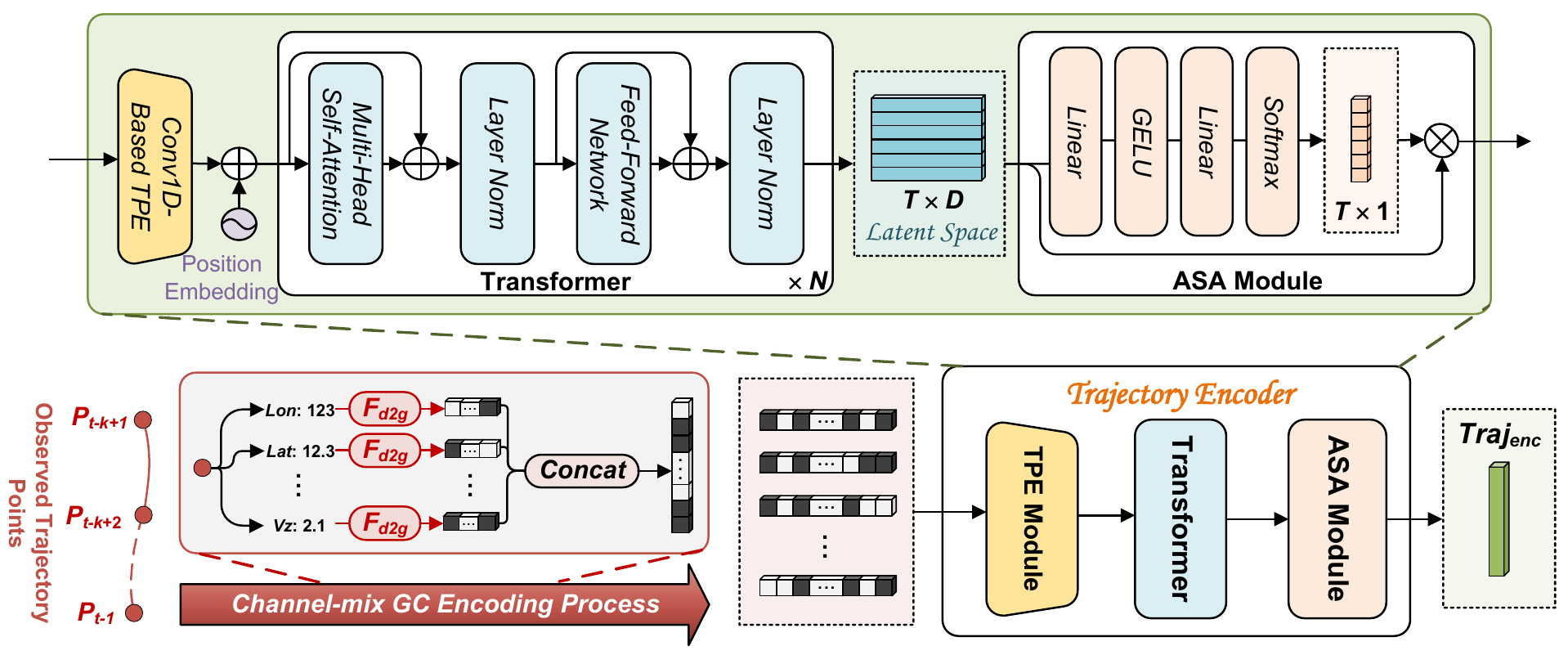}
	\caption{The detailed implementation of the channel-mix trajectory point embedding and trajectory encoder.} 
	\label{fig2}
\end{figure*}

To capture the informative spatial features of the trajectory points, 
the linear-based attributes embedding block is designed in \cite{9945661} to learn attribute embeddings in a channel-independent manner. 
However, the channel-independent approaches suffer from two limitations. i) Since strong spatial correlations are implied among the trajectory attributes, learning the attributes embedding separately might not fully consider the global features. 
ii) The parallel linear layers in the attributes embedding block burden the complexity of the model. 

To address the aforementioned limitations, in this work, a simple yet efficient Conv1D-based channel-mix trajectory point embedding module is proposed to project the GC representations of trajectory points into high-dimensional embedding space. 
As illustrated in Figure \ref{fig2}, for a trajectory point, a joint embedding is obtained by concatenating its attribute-wise GC representations, which is further fed into a Conv1D layer to learn trajectory point embedding. 
On the one hand, the channel-mix strategy retains the global correlations of the trajectory attributes. 
On the other hand, the Conv1D neural network is beneficial for learning the local features among the bits of the joint GC representation.  

The Transformer-based temporal modeling module is implemented by the stacked Transformer blocks. 
Compared to the RNN-based architectures, the Transformer is primarily implemented by the self-attention mechanism, which enables the model to learn the temporal correlations of the observations in a non-autoregressive manner. 
The ASA module performs a weighted sum operation in the temporal dimension, in which the attention weights of the trajectory points are generated as Figure \ref{fig2}. 
In this way, the trajectory-level embedding $\mathbf{Traj}_{enc}$ can be obtained from the trajectory encoder, which is expected to capture the temporal-spatial semantic representation of the observations.

\subsection{Horizon-Aware Context Generator} \label{sec3.5}
Conventional multi-horizon approaches usually perform the prediction in an autoregressive manner, i.e., predicting $p_{t+1}$ based on observation and generating the $p_{t+2}$ according to the $p_{t+1}$ sequentially. 
However, the autoregressive nature of these approaches degrades the efficiency in the training and inference process, especially in high temporal resolution conditions, which may be not suitable for real-time forecasting scenarios. 
Towards this gap, a horizon-aware context generator is innovatively designed to support multi-horizon predictions in a non-autoregressive manner by generating multi-horizon contexts directly. 

The architecture of the HACG is illustrated in Figure \ref{fig1}. 
Specifically, we conduct multi-horizon contexts by considering both the prior temporal specificities of the predicted horizons and trajectory-level embeddings of the observations. 
The inference rules of the proposed HACG can be shown below. 
Firstly, the predicted horizons are represented by a set of integer tokens, which are further encoded into corresponding one-hot vectors $\mathbf{H}$. 
Secondly, as shown in Eq. (\ref{eq41}), these one-hot vectors are projected into embedding space to learn the horizon embeddings $\mathbf{He}_{1:n} = \{\mathbf{he}_{1}, \mathbf{he}_{2}, ..., \mathbf{he}_{n}\}$ by horizon embedding module. 
\begin{equation} \label{eq41}
        \mathbf{He}_{1:n} = \operatorname{HorizonEmbedding}(\mathbf{H})
\end{equation}

In succession, the horizon embeddings and the trajectory-level embedding $\mathbf{Traj}_{enc}$ are concatenated to generate the context vector $\mathbf{hc}_{t+i}$ of horizon $i$. 
\begin{equation} \label{eq42}
        \mathbf{hc}_{t+i} = \operatorname{Concat}[\mathbf{Traj}_{enc},\mathbf{he}_{i}]
\end{equation}

Finally, the context vectors are further fed into an MLP block to perform high-dimensional projection and generate the final multi-horizon context representations $\mathbf{C}_{t+1:t+n}$. 
\begin{equation} \label{eq43}
        \mathbf{C}_{t+1:t+n} = \operatorname{MLP}([\mathbf{hc}_{t+1}, ..., \mathbf{hc}_{t+i}, ..., \mathbf{hc}_{t+n}])
\end{equation}

The core idea of the HACG is to leverage the trajectory-level embedding $\mathbf{Traj}_{enc}$ and horizon embeddings $\mathbf{He}_{1:n}$ to generate informative multi-horizon context representations. 
Specifically, the $\mathbf{Traj}_{enc}$ can be regarded as a high-dimensional representation that implies the global features of the observations, while the horizon embeddings $\mathbf{He}_{1:n}$ provide the different semantic representations for each predicted horizon.
Based on this assumption, the concatenation operation and MLP block are employed to fuse these vectors and generate the multi-horizon context representations. 
By integrating the prior horizon knowledge, the HACG is able to be aware of different horizons and generate the multi-horizon context representations directly via only one-pass inference.

\subsection{Differential-prompted Decoder} \label{sec3.6}
\begin{figure}[h]
	\centering
	\includegraphics[width=0.45 \textwidth]{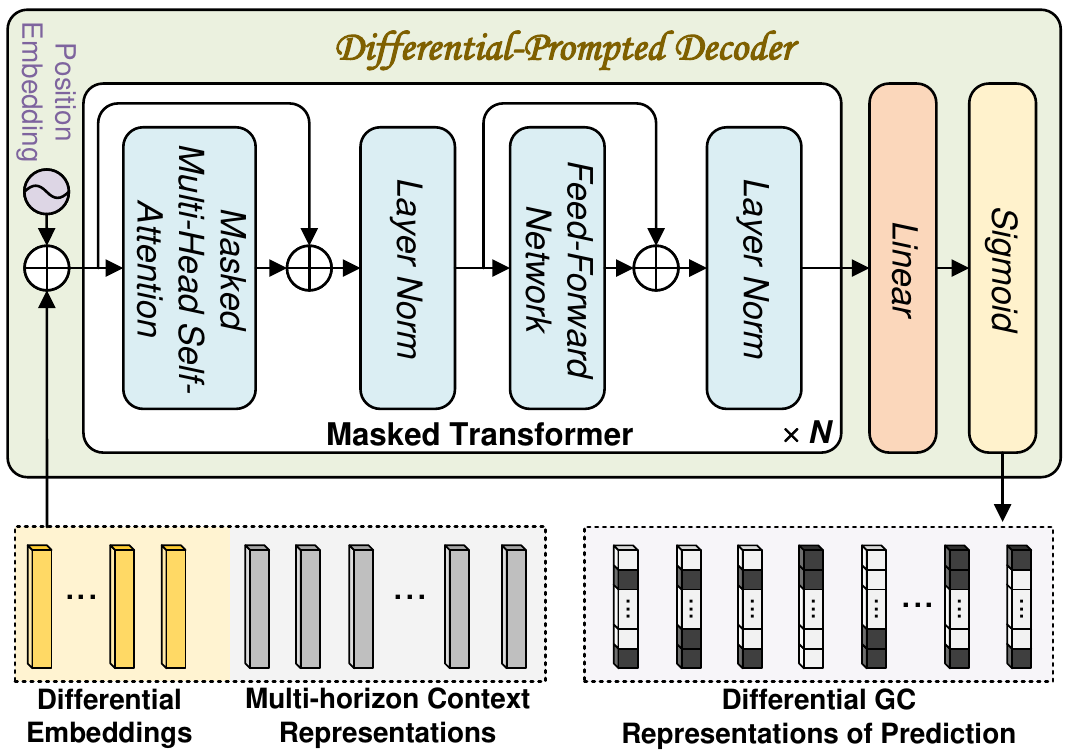}
	\caption{The detailed implementation of the Differential-prompted Decoder.} 
	\label{fig3}
\end{figure}

As described in Section \ref{sec3.2-1}, to suppress outliers caused by high-bit misclassifications, the differential prediction paradigm is introduced into the FlightBERT++ framework.
In practice, it is challenging to learn the transition patterns of differential sequence from the observations sequence because the differential operation may ignore some geographical and kinematical features of the trajectory attributes. 
To this end, as illustrated in Figure \ref{fig1} and Figure \ref{fig3}, a differential-prompted decoder is proposed to reduce the learning difficulty of the network by integrating the differential prompted mechanism. 
Specifically, the differential-prompted decoder consists of two modules, i.e., masked-Transformer, and predictor. 
Firstly, the differential sequence is calculated from the observed sequence and encoded into GC representations which can be consistent with the form of the outputs. 
Similar to the TPE module, these GC representations are further fed into a Conv1D-based embedding layer to learn the high-dimensional differential embeddings $\mathbf{D}_{t-k+2:t}$. 
Secondly, the learned differential embeddings $\mathbf{D}_{t-k+2:t}$ are concatenated with the multi-horizon context representations $\mathbf{C}_{t+1:t+n}$ along the temporal dimension as a prompt to learn the transition patterns of the differential sequence. 

In succession, the concatenated vectors further fed into the masked-Transformer module to build the inter- and intra-temporal correlations across observation and the multi-horizon predictions. 
The architecture of the masked-Transformer module is similar to the Transformer-based temporal modeling module except it employs the masked self-attention mechanism to ensure the temporal specificities of the sequence. 
Finally, the predictor is composed of a Linear layer and the Sigmoid activation, which is applied to predict the GC representations for the multi-horizons.

\subsection{Loss Function} \label{sec3.7}
Thanks to the nature of the GC representations, the proposed FlightBERT++ framework is formulated as the MBC paradigm. 
In this work, the BCE loss function is employed to optimize the network parameters in the training procedure. 
Specifically, let the $p, y$ are the ground truth and prediction of a trajectory point, 
$p_{a_j}^b$, $y_{a_j}^b$ are the GC representation of the $j^{th}, j \in [1,6]$, attribute in the ground truth and prediction, respectively. 
The calculation of the loss can be mathematically described as Eq. (\ref{eq6}) - Eq. (\ref{eq7}): 
\begin{equation} \label{eq6}
        \mathcal{L}(p,y) = \frac{1}{N} \sum_{n=1}^N \sum_{j=1}^J \mathcal{L}_{BCE}^{a_j}(p_{a_j}^b,y_{a_j}^b) 
\end{equation}
\begin{equation} \label{eq7}
        \begin{aligned}
                \mathcal{L}_{BCE}^{a_j}(p_{a_j}^b,y_{a_j}^b) = - \frac{1}{M} \sum_{i=1}^M (p_{a_j}^b[i] * log(y_{a_j}^b[i]) + \\ (1 - p_{a_j}^b[i]) * log(1 - y_{a_j}^b[i])) 
        \end{aligned}
\end{equation}

\noindent where the $N$ is the number of samples in the minibatch, while $M$ denotes the dimension of the GC representation of each attribute.

\section{Experimental Settings} \label{sec4}
\subsection{Dataset Preprocessing and Description} \label{sec4.1}
To validate the proposed framework, a real-world flight trajectory dataset was collected from an ATC system in China \cite{m2ats}. 
The dataset contains a total of 9 days of trajectory data with 20 seconds intervals from February 19 to 27, 2021, in which the range of the interested (ROI) longitude and latitude are $[94.616, 113.689]$ and $[19.305, 37.275]$, respectively. 
The key attributes of the flight trajectory are extracted from the raw data to support the experiments, including timestamps, call sign, longitude, latitude, altitude, and velocity in $x,y,z$ directions. 
The positional components (longitude, latitude, and altitude) are in degree with the WGS-84 coordinate system, while the velocity in $x,y,z$ dimensions are measured by $km/h$ with the Cartesian coordinates.

After the aforementioned preprocess, a total of 8643 flight trajectories in the dataset and split into train, validate, and test subsets. 
Specifically, the trajectory of the first 7 days is used to train the FTP models while $8^{th}$ and $9^{th}$ are for validation and testing, respectively. 
The aforementioned data splits are applied to all the experiments in this work to ensure fair comparison. 

\subsection{Comparison Baselines}
In this work, a total of 7 competitive approaches serve as baselines to validate the effectiveness of the proposed FlightBERT++ framework. 
Moreover, the baselines are divided into two groups to conduct comparisons by multi-horizon prediction styles. 
Group A performs iterative multi-horizon prediction, i.e., the model only predicts the result for the next time step and serves as a pseudo-observation for multi-horizon inference. 
Group B is the direct multi-horizon prediction approach that forecasts the results in a single inference process for multiple time steps. 
The detailed descriptions of the baselines are listed as follows.

\begin{itemize}
    \item (A1) LSTM: An FTP framework using LSTM neural networks \cite{ShiXPYZ18}. In this work, we employ a 4-layer LSTM network with 128 neurons to perform the FTP task. 
    \item (A2) Transformer: The Transformer block \cite{VaswaniSPUJGKP17} server as backbone network to build the FTP model. 
    The model is implemented by stacking 4 Transformer blocks with 128-dimensional hidden states and an FC layer. 
    \item (A3) Kalman-Filter: A typical model-driven flight state estimation algorithm based on historical observations and generally applied in target tracking scenarios. 
    In this work, the Kalman-Filter (KF) is employed to perform the FTP task \cite{1960A}. 
    \item (A4) FlightBERT: A novel FTP framework proposed in our previous work, which converts the FTP tasks into MBC paradigm and achieves desired performance \cite{9945661}. 
    \item (A5) WTFTP: A time-frequency analysis FTP framework using wavelet transform, achieves higher precision and robustness, particularly in high-maneuvering flight scenarios. 
    The model configuration is referred to the original WTFTP work, more details can be found in \cite{Zhang_2023}. 
    \item (B1) LSTM+Attention: A Seq2Seq architecture is designed to forecast the flight status across multiple future time steps directly. 
    In this experiment, an Attention-based encoder-decoder LSTM network \cite{BahdanauCB14} is adapted to the FTP task. 
    A 4-layer LSTM encoder, a 2-layer LSTM decoder, and a single-head attention module are employed to conduct the FTP model. 
    \item (B2) Transformer-Seq2Seq: In view of the Transformer achieving superior performance in sequential modeling tasks, a vanilla Seq2Seq Transformer architecture \cite{VaswaniSPUJGKP17} is employed to implement the FTP task. 
    In this model, the encoder and decoder are built by 4 and 2 Transformer blocks, with a 4-head multi-head attention mechanism, respectively.
\end{itemize}

\subsection{Experimental Configuration}
Based on ATC work, the quantization precision $0.001\degree$ (about 110 meters) is selected to adapt the transformation between GC representation and the decimals. 
In this resolution, the number of bits of the GC representation of trajectory attribute for inputs and differential values for outputs is listed as follows: 
\begin{itemize}
        \item Longitude and latitude: we use 18 and 16 bits to encode the real values (decimals) of longitude and latitude into GC representation for the inputs. 
        The number of the bits for longitude and latitude differential values are both set to 8 bits. 
        \item Altitude: the altitude is measured in $10 m$, and encoded into the GC with 11 bits, while its differential value encodes to GC representation with 8 bits. 
        \item Velocity: the velocity is measured in Km/h and all are encoded into the GC representation with 11 bits for inputs. 
        For the model outputs, the differential values of $Vx, Vy$ are represented by the 9-bit GC representation while the $Vz$ is encoded with a 6-bit GC representation. 
\end{itemize} 

Note that the highest bit of GC representations encoded from differential values indicates the sign of the real value. 
Based on the above configurations, the input of the proposed framework is a 78-dimensional vector while the output is a 48-dimensional vector. 
The embedding size of the trajectory attributes and horizons is set to 128. 
The number of the Transformer blocks in the trajectory encoder and differential-prompted decoder are set to 4 and 2, respectively. 
The number of hidden states of the Transformer blocks is set to 768. 
An attention operator with 4 heads is applied to all Transformer blocks in the proposed framework. 

For experiments A1, A2, B1, and B2, the Z-Score normalization algorithm is applied to process the value into [0,1] for longitude and latitude attributes due to the sparse specificities of their distributions. 
While the altitude and velocities in $x,y,z$ dimension are normalized into [0,1] using the Max-min algorithm. 
The MSE loss function is applied to optimize the above models in the training process. 

In the training phase, we use the latest 3-minute observations to predict the flight status of the future 5 minutes, i.e., predicting 15 trajectory points based on 9 observed trajectory points. 
The Adam optimizer with $10^{-4}$ initial learning rate is applied to train all the above deep learning-based models.

In this work, all the experiments are implemented with the open-source deep learning framework PyTorch 1.9.0. 
The models are trained on the server configured with Ubuntu 16.04 operating system, 8*NVIDIA GeForce RTX 2080 GPU, Intel(R) Core(TM) i7-7820X@3.6GHz CPU, and 128 GB memory. 

\subsection{Evaluation Metrics}
In this work, the Mean Absolute Error (MAE), Mean Absolute Percentage Error (MAPE), and Root Mean Squared Error (RMSE) are applied to evaluate the proposed methods and baselines, which are the common criteria for the TP tasks.  
The definitions of the MAE, MPAE, and RMSE are described as follows: 

\begin{equation} \label{eq12}
        MAE = \frac{1}{N} \frac{1}{h} \sum_{i=1} ^ N \sum_{j=1} ^ h   { \lvert a_{ij} - a^{\prime}_{ij}  \rvert }
\end{equation}

\begin{equation} \label{eq13}
        MAPE= \frac{1}{N} \frac{1}{h}  \sum_{i=1} ^ N \sum_{j=1} ^ h { \lvert \frac{a_{ij} - a^{\prime}_{ij}}{a_{ij}}  \rvert } \times 100 \% 
\end{equation}

\begin{equation} \label{eq14}
        RMSE = \sqrt{\frac{1}{N} \frac{1}{h} \sum_{i=1}^{N} \sum_{j=1} ^ h \left(a_{ij} - a^{\prime}_{ij} \right)^{2}}
\end{equation}

\noindent where $N$ is the number of the samples in the test set, $h$ represents the prediction horizon. 
$a, a^{\prime}$ are the real-value (decimals) of trajectory attributes in the ground truth and prediction, respectively. 

Although the longitude, latitude, and altitude (LLA) are measured using the above common metrics, it is not intuitive to evaluate the error separately 
due to the trajectory point being determined by LLA together in the three-dimensional (3D) airspace. 
To this end, the Mean Deviation Error (MDE) is proposed to evaluate the Euclidean distance ($km$) of the predictions and the actual trajectory points. 
Specifically, the LLA of both predictions and ground truth is projected to the earth-centered and earth-fixed (ECEF) coordinate system from the WGS-84 coordinate systems, 
and the deviation error in 3D airspace is measured to evaluate the model performance. 

\begin{equation} \label{eq15}
        MDE = \frac{1}{N} \frac{1}{h} \sum_{i=1} ^ N \sum_{j=1} ^ h \Phi( p_{ij} - p^{\prime}_{ij} ) 
\end{equation}

\noindent where $p, p^{\prime}$ are the transformed values of the ground truth and prediction in the ECEF coordinate system, respectively. 
$\Phi(\cdot)$ is the calculation function of Euclidean distance in the 3D airspace.

Furthermore, the Mean Time Costs (MTC) metric is proposed to evaluate the computational performance of the proposed framework and comparison baselines, defined as follows: 
\begin{equation} \label{eq16}
        MTC = \frac{1} {N} \sum_{i=1} ^ N time\_costs_{i}^{h} 
\end{equation}
\noindent where $N$ is the number of samples in the testing process, $time\_costs_{i}^{h}$ represents the time cost for $h$ prediction horizons of sample $i$. 
In this phase, the batch size of all evaluation models is set to 1 to ensure comparison fairness.  

% \usepackage{multirow}
\begin{table*}[htbp]
        \centering
        \caption{The experimental results of the proposed framework and baselines.} \label{tab1}
        \begin{tabular}{c|c|c|ccc|ccc|ccc|c}
        \hline\hline
        \multirow{2}{*}{\textbf{Style}} & \multirow{2}{*}{\textbf{Methods}} & \multirow{2}{*}{\textbf{Horizon}} & \multicolumn{3}{c|}{\textbf{MAE} $\downarrow$} & \multicolumn{3}{c|}{\textbf{MAPE (\%)} $\downarrow$} & \multicolumn{3}{c|}{\textbf{RMSE} $\downarrow$} & \multirow{2}{*}{\textbf{MDE} $\downarrow$} \\ \cline{4-12}
                                        &                                   &                                   & Lon        & Lat       & Alt      & Lon          & Lat         & Alt        & Lon        & Lat        & Alt      &                               \\ \hline
        \multirow{20}{*}{Iterative}     & \multirow{4}{*}{\makecell{LSTM\\(A1)}}             & 1                & 0.0054     & 0.0055    & 1.71     & 0.0050       & 0.0202      & 0.29       & 0.0093     & 0.0124     & 6.04     & 0.91                          \\
                                        &                                   & 3                                 & 0.0065     & 0.0065    & 3.01     & 0.0061       & 0.0238      & 0.50       & 0.0110     & 0.0151     & 7.63     & 1.08                          \\
                                        &                                   & 9                                 & 0.0127     & 0.0121    & 10.33    & 0.0118       & 0.0443      & 1.48       & 0.0239     & 0.0270     & 18.48    & 2.05                          \\
                                        &                                   & 15                                & 0.0214     & 0.0194    & 20.74    & 0.0199       & 0.0714      & 2.74       & 0.0436     & 0.0427     & 33.06    & 3.38                          \\ \cline{2-13} 
                                        & \multirow{4}{*}{\makecell{Transformer\\(A2)}}      & 1                & 0.0038     & 0.0040    & 1.59     & 0.0036       & 0.0146      & 0.26       & 0.0071     & 0.0133     & 8.81     & 0.65                          \\
                                        &                                   & 3                                 & 0.0073     & 0.0073    & 3.24     & 0.0068       & 0.0268      & 0.54       & 0.0127     & 0.0172     & 9.12     & 1.21                          \\
                                        &                                   & 9                                 & 0.0173     & 0.0172    & 8.68     & 0.0161       & 0.0628      & 1.37       & 0.0305     & 0.0325     & 17.19    & 2.85                          \\
                                        &                                   & 15                                & 0.0274     & 0.0270    & 13.87    & 0.0256       & 0.0986      & 2.08       & 0.0496     & 0.0501     & 26.25    & 4.50                          \\ \cline{2-13} 
                                        & \multirow{4}{*}{\makecell{Kalman-Filter\\(A3)}}    & 1                & 0.0067     & 0.0032    & 1.50     & 0.0062       & 0.0120      & 0.28       & 0.0054     & 0.0171     & 8.50     & 0.82                          \\
                                        &                                   & 3                                 & 0.0112     & 0.0059    & 2.97     & 0.0104       & 0.0220      & 0.54       & 0.0861     & 0.0272     & 11.48    & 1.41                          \\
                                        &                                   & 9                                 & 0.0281     & 0.0168    & 8.73     & 0.0261       & 0.0624      & 25.83      & 0.1847     & 0.0619     & 25.83    & 3.69                          \\
                                        &                                   & 15                                & 0.0494     & 0.0313    & 15.63    & 0.0459       & 0.1162      & 2.53       & 0.2896     & 0.1016     & 42.39    & 6.62                          \\ \cline{2-13} 
                                        & \multirow{4}{*}{\makecell{FlightBERT\\(A4)}}       & 1                & 0.0024     & 0.0021    & 1.20     & 0.0023       & 0.0077      & 0.23       & 0.0387     & 0.0325     & 12.04    & 0.44                          \\
                                        &                                   & 3                                 & 0.0039     & 0.0036    & 2.19     & 0.0035       & 0.0133      & 0.41       & 0.0486     & 0.0506     & 13.65    & 0.71                          \\
                                        &                                   & 9                                 & 0.0091     & 0.0086    & 6.20     & 0.0085       & 0.0317      & 1.09       & 0.0608     & 0.0679     & 22.45    & 1.60                          \\
                                        &                                   & 15                                & 0.0159     & 0.0148    & 10.80    & 0.0148       & 0.0549      & 1.84       & 0.0742     & 0.0794     & 32.76    & 2.71                          \\ \cline{2-13}
                                        & \multirow{4}{*}{\makecell{WTFTP\\(A5)}}            & 1                & 0.0021     & 0.0021    & 1.10     & 0.0020       & 0.0077      & 0.18       & 0.0049     & 0.0105     & 6.29     & 0.35                          \\
                                        &                                   & 3                                 & 0.0034     & 0.0033    & 2.33     & 0.0032       & 0.0124      & 0.39       & 0.0077     & 0.0134     & 7.63     & 0.56                          \\
                                        &                                   & 9                                 & 0.0109     & 0.0099    & 7.68     & 0.0102       & 0.0367      & 1.21       & 0.0233     & 0.0265     & 16.76    & 1.71                          \\
                                        &                                   & 15                                & 0.0211     & 0.0187    & 14.40    & 0.0197       & 0.0693      & 2.19       & 0.0429     & 0.0432     & 27.80    & 3.28                          \\ \hline
        \multirow{12}{*}{Direct}        & \multirow{4}{*}{\makecell{LSTM+Attention\\(B1)}}   & 1                & 0.0064     & 0.0068    & 1.98     & 0.0059       & 0.0249      & 0.30       & 0.0109     & 0.0138     & \bf{5.96}    & 1.09                      \\
                                        &                                   & 3                                 & 0.0058     & 0.0060    & 2.75     & 0.0054       & 0.0221      & 0.43       & 0.0101     & 0.0144     & \bf{7.36}    & 0.98                      \\
                                        &                                   & 9                                 & 0.0082     & 0.0083    & 6.07     & 0.0076       & 0.0305      & 0.94       & 0.0178     & 0.0221     & \bf{14.03}   & 1.37                      \\
                                        &                                   & 15                                & 0.0125     & 0.0122    & 9.05     & 0.0116       & 0.0450      & 1.38       & 0.0299     & 0.0327     & \bf{20.01}   & 2.04                      \\ \cline{2-13} 
                                        & \multirow{4}{*}{\makecell{Transformer-Seq2Seq\\(B2)}}  & 1            & 0.0041     & 0.0040    & 2.05     & 0.0038       & 0.0146      & 0.31       & 0.0063     & 0.0095     & 6.33     & 0.67                          \\
                                        &                                   & 3                                 & 0.0044     & 0.0045    & 2.86     & 0.0041       & 0.0165      & 0.45       & 0.0076     & 0.0122     & 7.71     & 0.74                          \\
                                        &                                   & 9                                 & 0.0078     & 0.0076    & 6.08     & 0.0073       & 0.0279      & 0.95       & 0.0174     & \bf{0.0204}& 14.27    & 1.28                          \\
                                        &                                   & 15                                & 0.0124     & 0.0117    & 9.15     & 0.0116       & 0.0434      & 1.41       & 0.0303     & 0.0312     & 20.31    & 2.01                          \\ \cline{2-13} 
                                        & \multirow{4}{*}{\makecell{FlightBERT++\\(AAAI version)}}     & 1      & 0.0017     & 0.0017    & 1.15     & 0.0016       & 0.0066      & 0.20       & 0.0037     & 0.0115     & 12.07    & 0.31                          \\
                                        &                                   & 3                                 & 0.0031     & 0.0031    & 2.23     & 0.0029       & 0.0117      & 0.41       & 0.0067     & 0.0131     & 12.46    & 0.55                          \\
                                        &                                   & 9                                 & 0.0076     & 0.0074    & 5.30     & 0.0070       & 0.0277      & 0.96       & 0.0172     & 0.0232     & 17.92    & 1.29                          \\
                                        &                                   & 15                                & 0.0124     & 0.0117    & 7.43     & 0.0109       & 0.0425      & 1.37       & 0.0265     & 0.0326     & 22.89    & 1.97                          \\ \cline{2-13}
                                        & \multirow{4}{*}{\makecell{FlightBERT++\\(Proposed)}}     & 1          & \bf{0.0012}     & \bf{0.0012}    & \bf{0.93}     & \bf{0.0011}       & \bf{0.0046}      & \bf{0.17}     & \bf{0.0031}     & \bf{0.0071}     & 8.86   & \bf{0.22}                          \\
                                        &                                   & 3                                 & \bf{0.0023}     & \bf{0.0023}    & \bf{1.92}     & \bf{0.0022}       & \bf{0.0086}      & \bf{0.35}     & \bf{0.0059}     & \bf{0.0115}     & 10.05  & \bf{0.39}                          \\
                                        &                                   & 9                                 & \bf{0.0058}     & \bf{0.0058}    & \bf{4.69}     & \bf{0.0054}       & \bf{0.0215}      & \bf{0.84}     & \bf{0.0156}     & 0.0213          & 15.58  & \bf{0.97}                          \\
                                        &                                   & 15                                & \bf{0.0091}     & \bf{0.0090}    & \bf{6.59}     & \bf{0.0085}       & \bf{0.0336}      & \bf{1.19}     & \bf{0.0239}     & \bf{0.0298}     & 20.25  & \bf{1.54}                         \\ \hline\hline
        \end{tabular}
\end{table*}

\section{Result and Discussions} \label{sec5}
\subsection{Results and Quantitative Analysis} \label{sec5.1}
\subsubsection{Overall Performance of FTP} \label{sec5.1.1}
\begin{figure}[tbp]
	\centering
	\includegraphics[width=0.45 \textwidth]{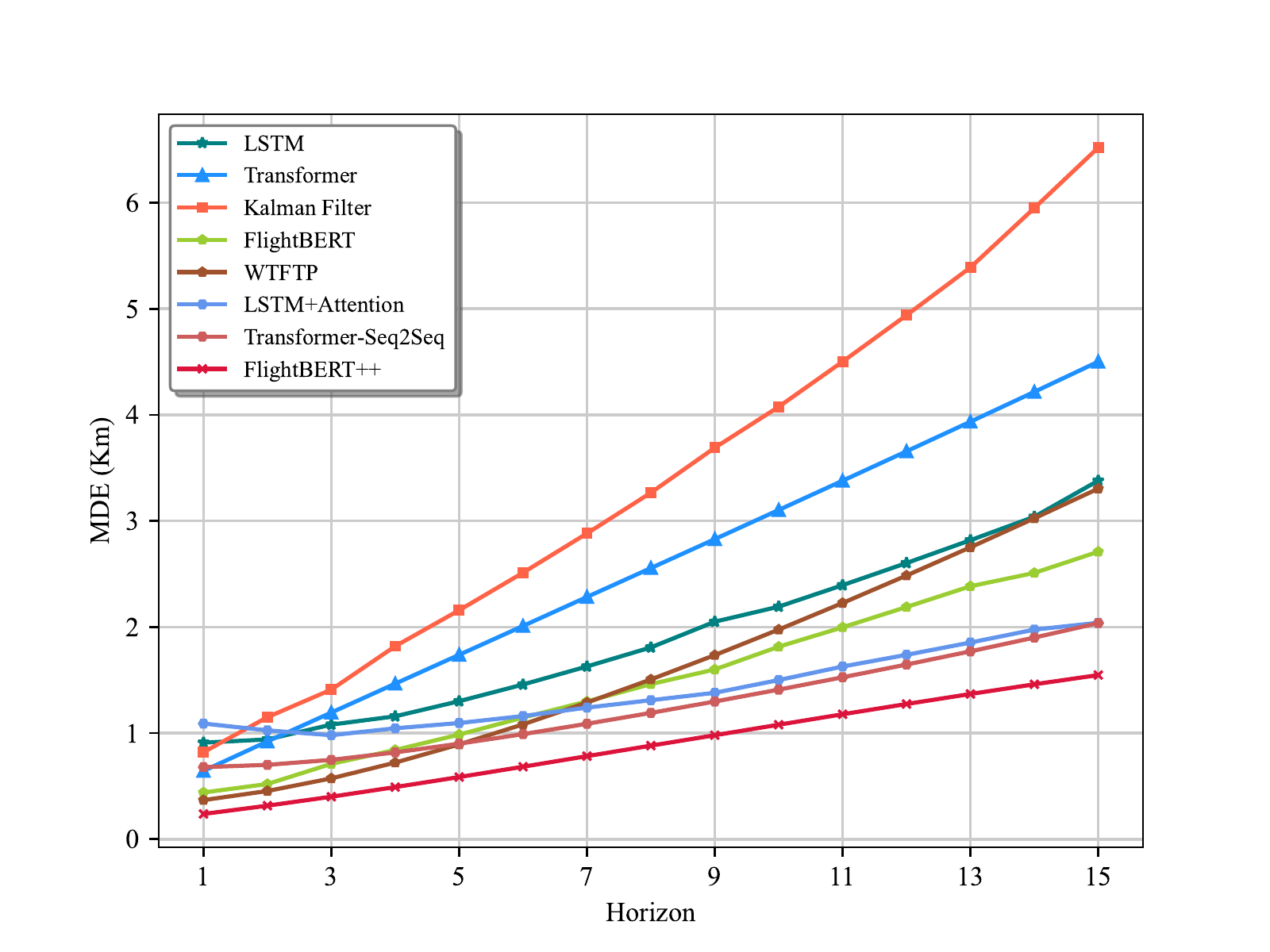}
	\caption{The comparison of the MDE curve over the 1 to 15 prediction horizons for different models.} 
	\label{fig_mde}
\end{figure}

Table \ref{tab1} reports the overall performance of the proposed framework and comparison baselines. 
To investigate the robustness of these models with the horizons increase, the experimental results are divided into four (1, 3, 9, 15) different horizons, corresponding to 20 seconds, 1, 3, and 5 minutes trajectories in the future. 
It is demonstrated that the proposed FlightBERT++ framework achieves significant performance improvements against FlightBERT and outperforms other baselines in the MAE and MAPE metrics across all predicted attributes. 
Thanks to the proposed GC representation and the reduction of the bits by the differential prediction paradigm, the outliers caused by the high-bit misclassifications are effectively eliminated, leading to substantial reductions in RMSE for longitude, latitude, and altitude within the proposed framework. 
In addition, benefiting from the dedicated network design of the proposed framework, the error accumulation of the multi-horizon predictions is significantly decreased compared to competitive baselines. 
Furthermore, the MDE curve of the proposed FlightBERT++ and comparison baselines are illustrated in Figure \ref{fig_mde}. 
It can be found that the proposed FlightBERT++ shows the lowest MDE accumulation with the increase of the prediction horizon. 

Notably, compared to the AAAI conference version, the FlightBERT++ in this paper achieves over 29.0\% MDE reduction in 1 and 3 prediction horizons, as well as 24.8\% and 21.8\% MDE reduction in 9 and 15 prediction horizons. 
These experimental results further demonstrate the effectiveness of the proposed new GC representation in the MBC-based FTP task. 

% KF 
For the iterative multi-horizon prediction approaches, the KF-based model-driven approach suffers from a huge performance degradation with the increasing of the prediction horizon. 
The primary reason is that the KF-based FTP approach needs to update the parameters of the state equation dynamically to correct the prediction errors. 
However, the loss of real-time observation updates procedure leads to serious error accumulation in the multi-horizon prediction process. 
Compared to the KF-based approach, the data-driven based approaches achieved better performance across all prediction horizons. 
The Transformer-based model achieves the desired performance of the 1- and 3-horizon predictions but fails to output more accurate forecasting in the 9- and 15-horizon procedures. 
In contrast, the performance of the LSTM is slightly better than Transformer-based models in long-term horizons but loses the precision of the output in 1 horizon. 
It can be attributed that the Transformer-based models are more sensitive to prediction errors of pseudo-observations, especially in large error accumulation conditions, 
while the LSTM-based model makes it hard to fit complex flight patterns of the diverse flight phases for better performance. 
Benefiting from the capacity of the BE representation, the FlightBERT obtains better MAE performance in all prediction horizons. 
However, the RMSE of the FlightBERT is higher than other data-driven baselines due to the outliers caused by high-bit misclassification in the prediction. 
While the WTFTP demonstrates superior performance within 3 prediction horizons compared to all baselines, it suffers from heavy error accumulation during the iterative multi-horizon inference process. 

The direct multi-horizon prediction baselines outperform iterative multi-horizon prediction approaches for long-term horizon predictions, particularly in horizons 9 and 15. 
However, the LSTM+Attention model suffers from the performance degradation of the short-term horizons (horizon 1) across both MAE and RMSE metrics. 
This result demonstrates that the LSTM+Attention approach primarily captures the global trajectory trends and dependencies over long-term horizons, but it falls short in effectively learning local and short-term fine-grained features. 
In contrast, the Transformer-Seq2Seq model achieves comparable results across both short and long prediction horizons, indicating its capability to effectively capture both global and local trajectory transition patterns.   

In summary, the direct multi-horizon prediction approaches are superior to the iterative approaches in long-term prediction steps since the global dependencies are learned from the training process. 
Thanks to the robust temporal-spatial dependencies obtained by the trajectory encoder and differential-promoted decoder, the FlightBERT++ framework achieves expected performance in both short- and long-term prediction horizons, 
which further demonstrates the effectiveness of the network design.

\subsubsection{Computational Performance Evaluation} 
\begin{table}[htbp]
        \caption{Comparison of the computational performance.} \label{tab_cp} 
        \centering
        \setlength\tabcolsep{14 pt} 
        \begin{tabular}{ccc}
        \hline \hline
        \textbf{Methods}      & \textbf{Parameters   (M)} & \textbf{MTC   (ms)} \\ \hline
        LSTM                  & 0.55                      & 48.96               \\
        Transformer           & 0.42                      & 63.18               \\
        Kalman-Filter         & --                        & 0.64                \\
        FlightBERT            & 56.85                     & 201.41              \\
        WTFTP                 & 0.22                      & 30.97               \\
        LSTM+Attention        & 0.90                      & 14.43               \\
        Transformer-Seq2Seq   & 0.76                      & 33.53               \\
        FlightBERT++          & 29.96                     & 6.95                \\ \hline \hline
        \end{tabular}
\end{table}

The computational performance and the size of the model parameters are presented in Table \ref{tab_cp}. 
To compare the prediction efficiency in multi-horizon settings, the number of the prediction horizon ($h$ in the Eq. (\ref{eq16})) is set to 15 for both FlightBERT++ and baseline models. 

As can be seen from the results, it is evident that the KF-based model achieves faster prediction speed due to its lower computational complexity and few parameters. 
Among the deep learning-based models, the direct multi-horizon approaches demonstrate substantial improvements in computational performance compared to the iterative multi-horizon approaches.  
In addition, it is worth noting that the model size of FlightBERT and FlightBERT++ is larger than comparison models. 
It can be attributed that the BE representations extend the dimension of the inputs and enable us to dedicatedly design the sophisticated neural architecture to capture the flight transition patterns. 
Moreover, the proposed FlightBERT++ harvests the fastest computational speed among deep learning based models, even with a tens larger model size than other baselines. 
This significant improvement in computational efficiency is primarily attributed to the design of the HACG module, enabling the FlightBERT++ to perform multi-horizon prediction in a non-autoregressive manner. 
In summary, the experimental results demonstrate that FlightBERT++ exhibits higher computational efficiency, making it well-suited for supporting real-time forecasting in ATC environments. 

Compared with the conference version \cite{guo2023flightbert++}, we streamline the network of the FlightBERT++ by setting the number of Transformer layers to 2 in the DPD, thereby reducing the number of parameters and computational complexity.
However, it is worth noting that a slight increase in MTC is observed (6.95 ms v.s. 6.81 ms) in this version due to the additional encoding process of the GC representation.

\subsection{Visualization and Qualitative Analysis} \label{sec5.2}
\begin{figure*}[!htbp] 
        \centering
        \subfloat[Descending]{\includegraphics[width=2.8in]{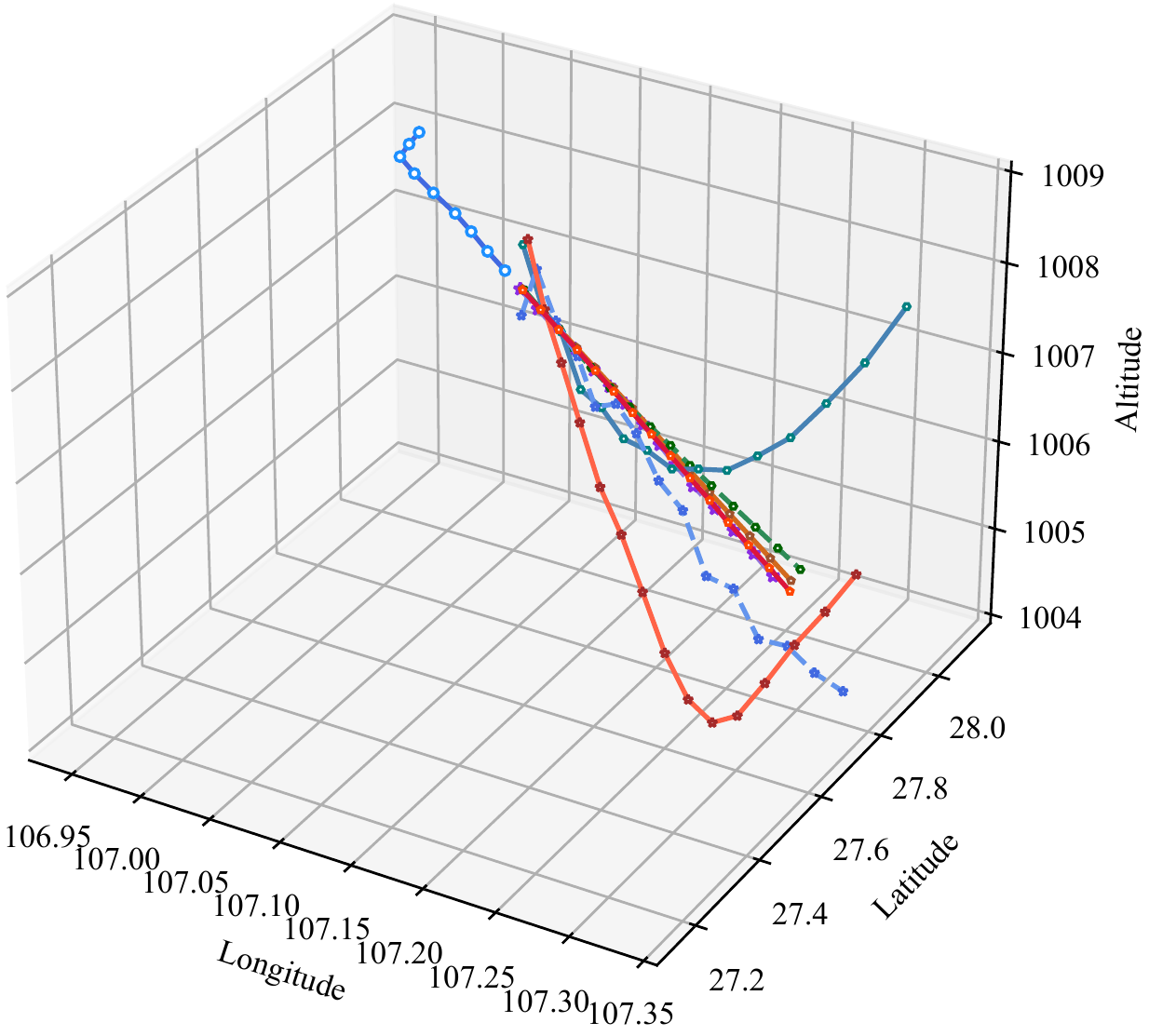} \label{fig_vis1} } \hspace{0.2 in}
        \subfloat[Climbing]{\includegraphics[width=2.8in]{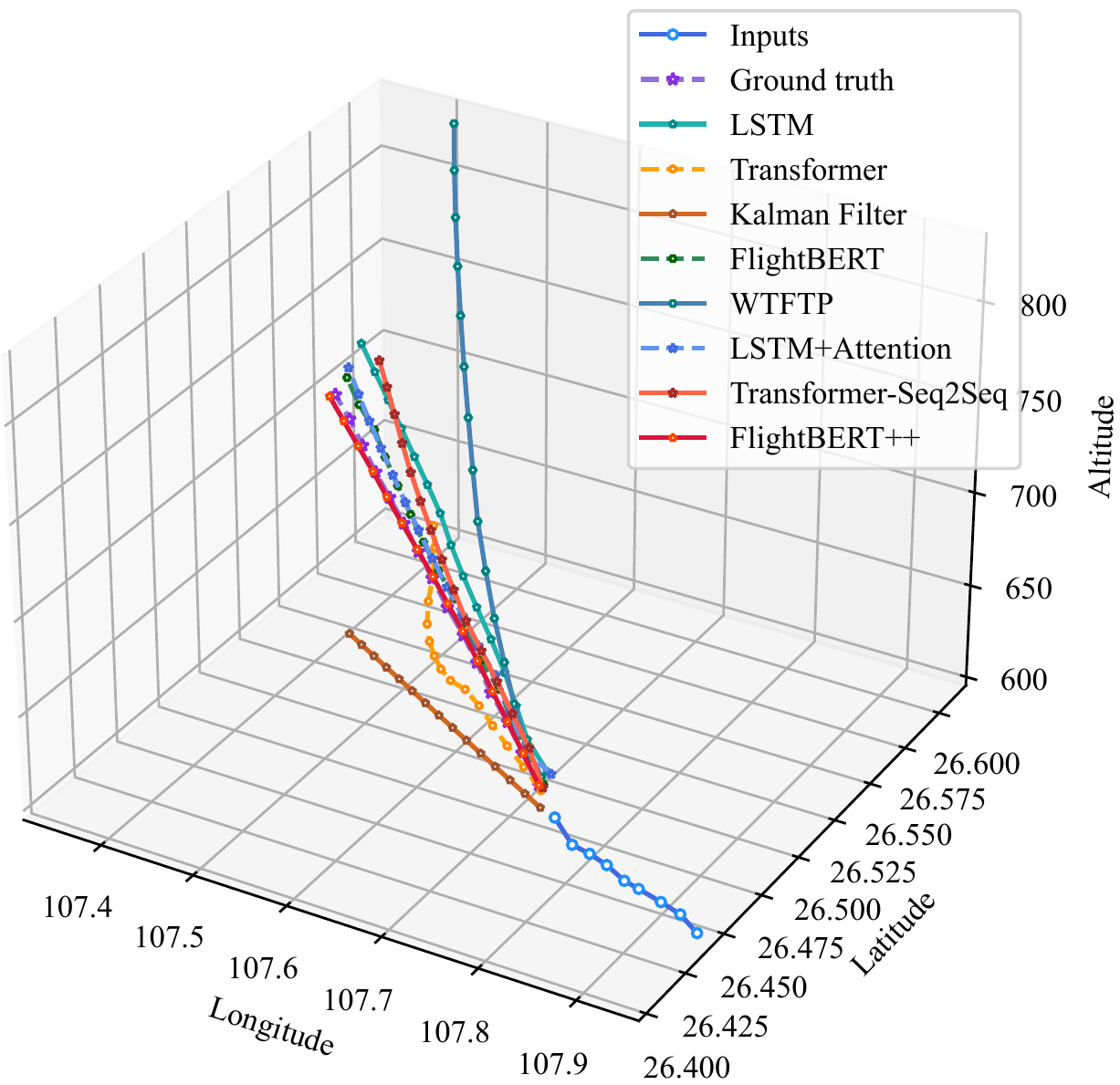} \label{fig_vis2}} \hspace{0.3in}

        \subfloat[En-route]{\includegraphics[width=2.8in]{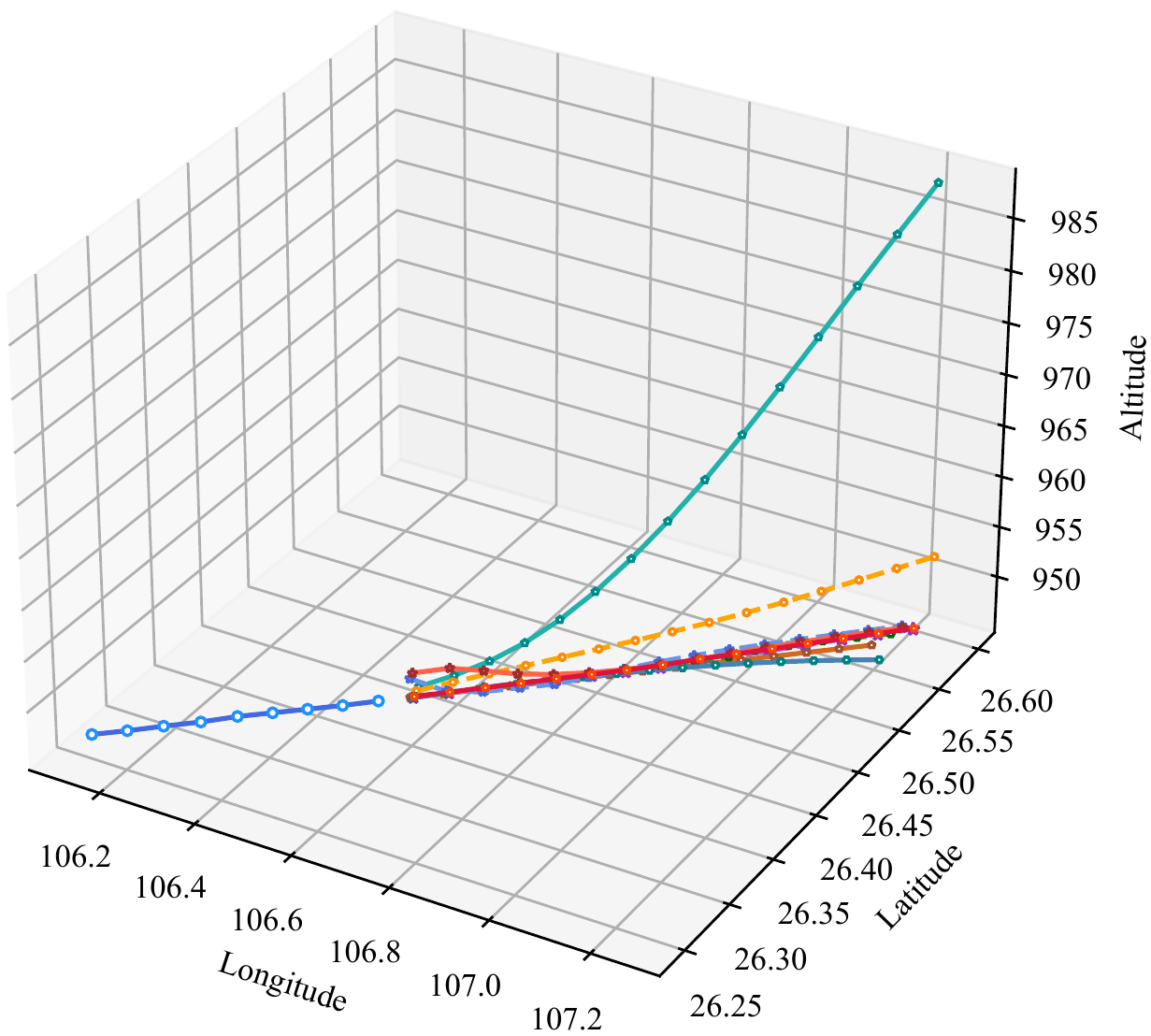} \label{fig_vis3}} \hspace{0.3in}
        \subfloat[Turn]{\includegraphics[width=2.8in]{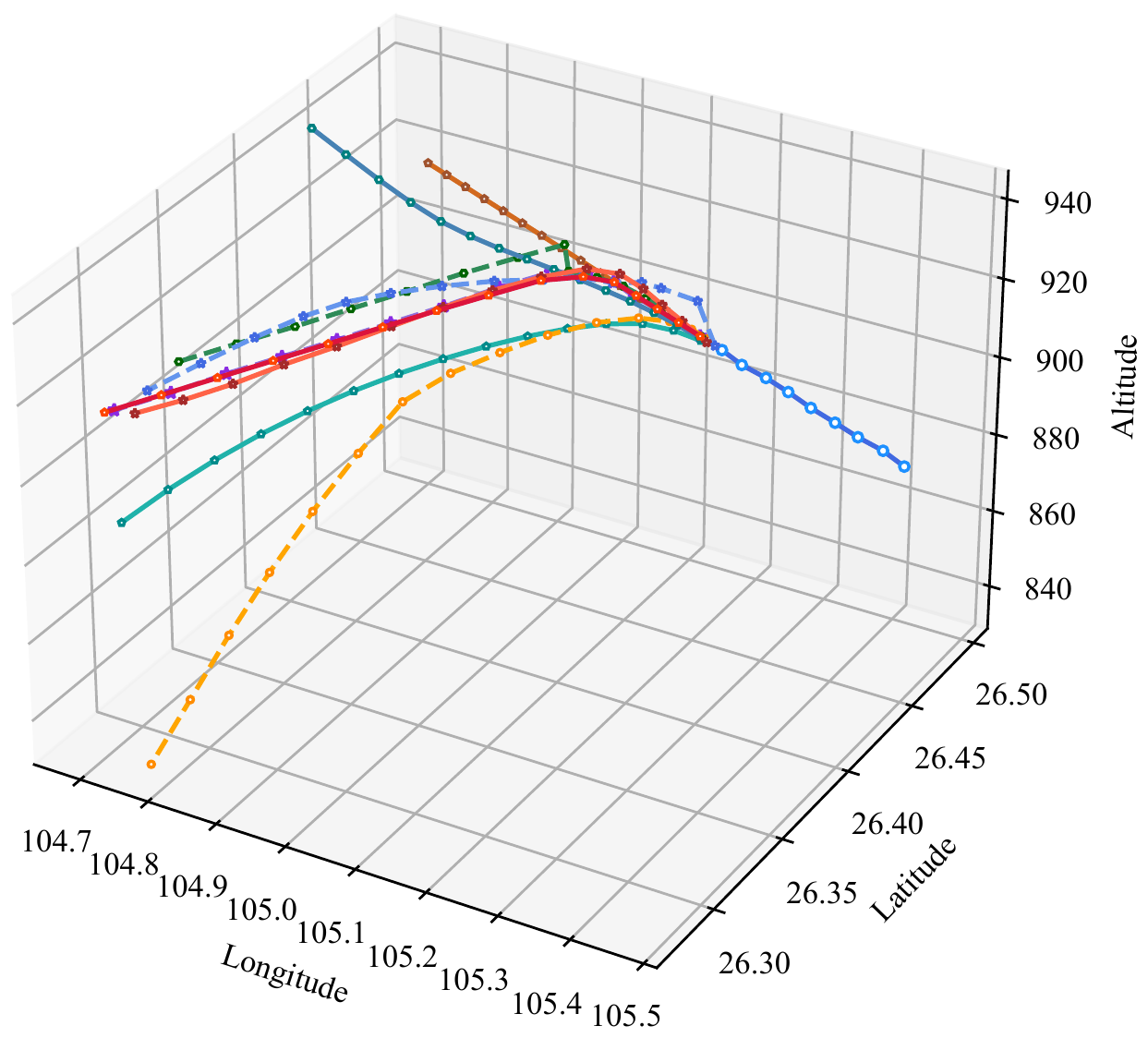} \label{fig_vis4}}

        \subfloat[Climbing and turn right]{\includegraphics[width=2.8in]{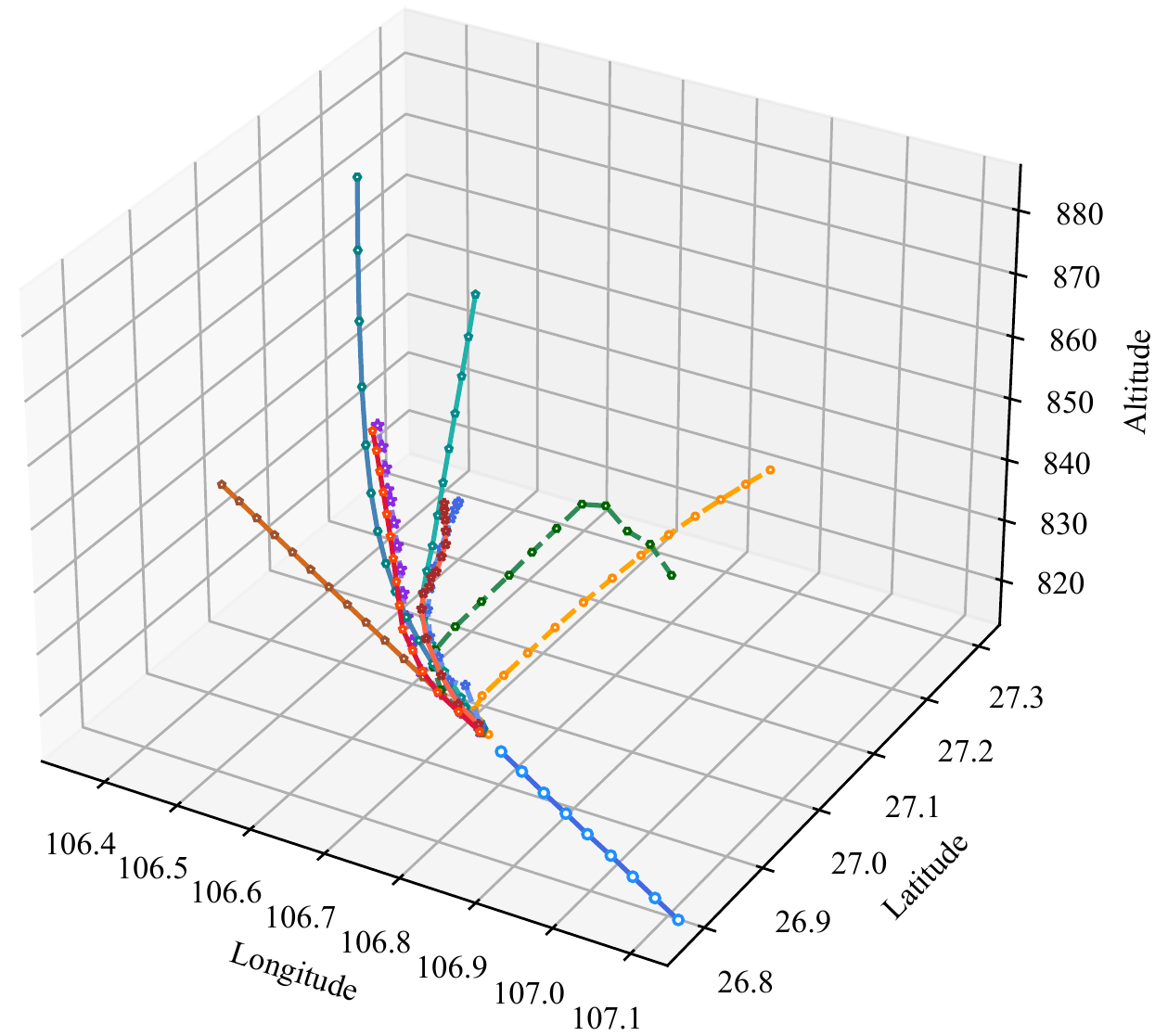} \label{fig_vis5} } \hspace{0.3in}
        \subfloat[Descending and maintain]{\includegraphics[width=2.8in]{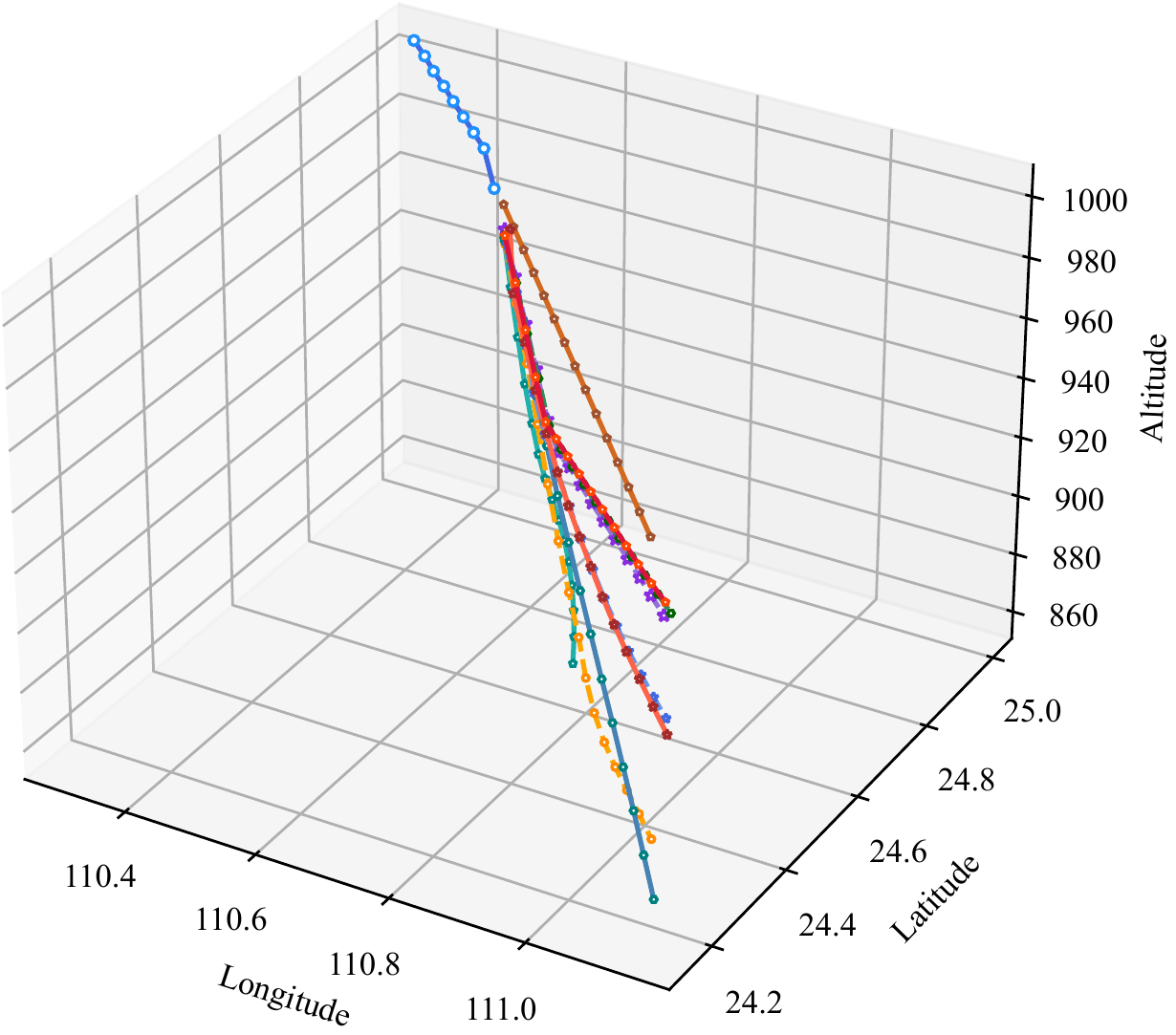} \label{fig_vis6}} 
        \caption{Visualization of the trajectory prediction results of selective flight scenarios, in which the altitude is measured in 10 $m$. Note that certain inaccurate predictions made by baselines are removed from
        figures to enhance readability.} \label{fig_vis}
\end{figure*}

In this section, to better understand the learned flight transition patterns and qualitatively analyze the performance of different approaches, a total of six typical flight scenarios in the test set are selected to visualize the prediction results.   
The visualization of selective samples is shown in Figure \ref{fig_vis}, including common flight scenarios (descending, climbing, and en-route) and complex flight patterns (turn, climbing and turn right, descending and maintain). 
Each sample is visualized in a 3D space with 9 observation trajectory points (inputs), ground truth, and 15 predicted trajectory points generated by the proposed FlightBERT++ framework and baseline models. 

As shown in Figure \ref{fig_vis}, the FlightBERT++ achieves superior performance over comparison baselines for both common flight scenarios and complex flight patterns. 
It is observed that the proposed model also exhibits the ability to estimate the flight intents in future horizons (Figure \ref{fig_vis4}, \ref{fig_vis5}, and \ref{fig_vis6}). 
This indicates that the FlightBERT++ not only captures the flight dynamics from the observations but also learns typical flight patterns, such as fixed waypoints of turns or descents, from a substantial amount of historical trajectories in training samples. 
Moreover, the FlightBERT++ shows lower error accumulation during the multi-horizon prediction process, indicating significant reliability over comparison approaches. 
Considering the enhancement of downstream tasks, such as conflict detection and airspace planning, 
the FlightBERT++ can be regarded as a powerful tool to improve the overall efficiency and safety of ATC operations.

For the LSTM+Attention and Transformer-Seq2Seq models, visualizations reveal that it is able to capture future flight trends for certain scenes. 
However, the LSTM+Attention model exhibits relatively larger prediction errors within 1 to 2 horizons (e.g., Figure \ref{fig_vis2}, \ref{fig_vis3}, and \ref{fig_vis4}), as that of the findings in the quantitative analysis. 
One possible reason is that the decoder of the LSTM+Attention model generates the 1 horizon predictions using a fixed $<\!\!sos\!\!>$ token, which brings a significant challenge in generating diverse and high-accurate results. 

Among the iterative multi-horizon prediction approaches, it is observed that the error accumulation is enlarged with the increase of the prediction horizon. 
These models are able to predict the flight status accurately within 1 to 3 horizons but suffer from larger errors in longer horizons. 
Notably, the LSTM and Transformer models exhibit significant error accumulation, particularly in the altitude dimension (e.g., Figure \ref{fig_vis3}, \ref{fig_vis4}, and \ref{fig_vis6}). 
The Kalman filter-based method can achieve satisfactory prediction performance only in simple or transition patterns with linear variation flight scenarios (e.g., Figure \ref{fig_vis1}, \ref{fig_vis2}, and \ref{fig_vis3}). 
In addition, the FlightBERT model outperforms other iterative multi-horizon prediction approaches in most flight scenarios. 
However, the ability of flight trends perception is limited by the single-horizon prediction training strategy, making it challenging to precisely estimate future flight intents in complex or maneuvering flight phase (e.g., Figure \ref{fig_vis4}, \ref{fig_vis5}).

\begin{table*}[htbp]
        \centering
        \setlength\tabcolsep{7 pt} 
        \caption{The experimental results of the ablation study.} \label{tab_ab}
        \begin{tabular}{c|c|c|ccc|ccc|ccc|c}
        \hline \hline
        \multirow{2}{*}{\textbf{Exp.}} & \multirow{2}{*}{\textbf{Models}}    & \multirow{2}{*}{\textbf{Horizon}} & \multicolumn{3}{c|}{\textbf{MAE}}    & \multicolumn{3}{c|}{\textbf{MAPE (\%)}}     & \multicolumn{3}{c|}{\textbf{RMSE}}  & \multirow{2}{*}{\textbf{MDE}} \\ \cline{4-12}
                                       &                                      &               & Lon     & Lat    & Alt   & Lon    & Lat    & Alt    & Lon    & Lat    & Alt    &                               \\ \hline
        \multirow{4}{*}{\textbf{C1}}   & \multirow{4}{*}{\makecell{FlightBERT++\\(using BE representation)}} & 1             & 0.0017    & 0.0017    & 1.14    & 0.0016    & 0.0062    & 0.22    & 0.0038    & 0.0074    & 9.06     & 0.31           \\
                                       &                                                                     & 3             & 0.0032    & 0.0031    & 2.32    & 0.0030    & 0.0115    & 0.43    & 0.0070    & 0.0122    & 10.91    & 0.55           \\     
                                       &                                                                     & 9             & 0.0076    & 0.0074    & 5.60    & 0.0071    & 0.0273    & 1.04    & 0.0172    & 0.0228    & 17.65    & 1.29           \\
                                       &                                                                     & 15            & 0.0117    & 0.0112    & 8.04    & 0.0109    & 0.0417    & 1.51    & 0.0262    & 0.0319    & 23.70    & 1.96           \\ \hline
        \multirow{4}{*}{\textbf{C2}}   & \multirow{4}{*}{\makecell{FlightBERT++\\(w/o differential prompted\\mechanism)}} & 1& 0.0014    & 0.0015    & 1.10    & 0.0013    & 0.0056    & 0.19    & 0.0039    & 0.0115    & 12.07    & 0.26           \\
                                       &                                                                     & 3             & 0.0027    & 0.0026    & 2.23    & 0.0025    & 0.0099    & 0.39    & 0.0078    & 0.0131    & 12.57    & 0.46           \\
                                       &                                                                     & 9             & 0.0072    & 0.0069    & 6.15    & 0.0067    & 0.0256    & 1.04    & 0.0216    & 0.0244    & 19.73    & 1.20           \\
                                       &                                                                     & 15            & 0.0126    & 0.0118    & 9.96    & 0.0117    & 0.0439    & 1.65    & 0.0369    & 0.0378    & 27.57    & 2.07           \\ \hline
        \multirow{4}{*}{\textbf{C3}}   & \multirow{4}{*}{\makecell{FlightBERT++\\(w/o ASA module)}}          & 1             & 0.0013    & 0.0013    & 1.01    & 0.0012    & 0.0051    & 0.18    & 0.0032    & 0.0114    & 11.98    & 0.23            \\
                                       &                                                                     & 3             & 0.0024    & 0.0024    & 1.96    & 0.0022    & 0.0089    & \bf{0.35}    & 0.0060    & 0.0129    & 12.15    & 0.41           \\
                                       &                                                                     & 9             & 0.0060    & 0.0059    & 4.66    & 0.0056    & 0.0221    & \bf{0.83}    & 0.0160    & 0.0225    & 16.74    & 1.01           \\
                                       &                                                                     & 15            & 0.0095    & 0.0094    & 6.43    & 0.0088    & 0.0348    & \bf{1.16}    & 0.0248    & 0.0318    & 20.77    & 1.58           \\ \hline
        \multirow{4}{*}{\textbf{C4}}   & \multirow{4}{*}{\makecell{FlightBERT++\\(using Linear-based TPE)}}  & 1             & 0.0013    & 0.0013    & 0.99    & 0.0012    & 0.0050    & 0.17    & \bf{0.0031}    & 0.0113    & 11.96    & 0.23           \\
                                       &                                                                     & 3             & 0.0023    & 0.0023    & 1.94    & \bf{0.0022}    & 0.0088    & 0.53    & \bf{0.0058}    & 0.0126    & 12.10    & 0.40           \\
                                       &                                                                     & 9             & 0.0059    & \bf{0.0058}    & \bf{4.41}& 0.0055    & 0.0217    & 0.83    & \bf{0.0156}    & 0.0216    & 16.63    & 0.99           \\
                                       &                                                                     & 15            & 0.0093    & 0.0093         & \bf{6.45} & 0.0087   & 0.0344    & 1.18    & 0.0242    & 0.0306    & 20.70    & 1.56           \\ \hline 
        \multirow{4}{*}{\textbf{--}}   & \multirow{4}{*}{\makecell{FlightBERT++\\(Proposed)}}                & 1             & \bf{0.0012}     & \bf{0.0012}    & \bf{0.93}     & \bf{0.0011}       & \bf{0.0046}      & \bf{0.17}     & \bf{0.0031}     & \bf{0.0071}     & \bf{8.86 }  & \bf{0.22}                          \\
                                       &                                                                     & 3             & \bf{0.0023}     & \bf{0.0023}    & \bf{1.92}     & \bf{0.0022}       & \bf{0.0086}      & \bf{0.35}     & 0.0059          & \bf{0.0115}     & \bf{10.05}  & \bf{0.39}                          \\
                                       &                                                                     & 9             & \bf{0.0058}     & \bf{0.0058}    & 4.69          & \bf{0.0054}       & \bf{0.0215}      & 0.84          & \bf{0.0156}     & \bf{0.0213}     & \bf{15.58}  & \bf{0.97}                          \\
                                       &                                                                     & 15            & \bf{0.0091}     & \bf{0.0090}    & 6.59          & \bf{0.0085}       & \bf{0.0336}      & 1.19          & \bf{0.0239}     & \bf{0.0298}     & \bf{20.25}  & \bf{1.54}                         \\ \hline\hline
        \end{tabular}
\end{table*}
\subsection{Ablation study} \label{sec5.3}
The efficiency and effectiveness of the proposed FlightBERT++ framework are demonstrated in the above sections by the qualitative and quantitative evaluations. 
In this section, to confirm the contributions of the designed components in the FlightBERT++ framework, an additional group C experiment is introduced to conduct the ablation study. 
Specifically, the BE representation is applied to construct the FlightBERT++ framework to evaluate the effectiveness of the proposed GC representation in experiment C1. 
The differential prompt mechanism is removed from the DPD in experiment C2, while a naive sum operator is applied to replace the ASA module in the trajectory encoder in experiment C3. 
Furthermore, in experiment C4, to validate the effectiveness of the proposed Conv1D-based channel-mix TPE module, a set of linear layers \cite{9945661} is employed to alternate the Conv1D layer to build a channel-independent TPE (Linear-based TPE) module in the FlightBERT++.  
In the above experiments, expected the validated modules, other experimental settings remain consistent with those outlined in Section \ref{sec4} to ensure experimental fairness and comparability.

The experimental results are presented in Table \ref{tab_ab}, in which the w/o represents the FlightBERT++ implemented without specific modules. 
It can be found from the results that all the designed components make expected contributions to the FlightBERT++ framework. 
In experiment C1, it is observed that significant performance improvement is obtained by using the proposed GC representation compared to the BE representation. 
Furthermore, in experiment C2, it can be found that the differential prompted mechanism is critical to the proposed FlightBERT++, especially in the long-horizon prediction conditions. 
On the one hand, the differential prediction paradigm of the proposed FlightBERT++ may lose some geographical and kinematical features of the flight trajectory, which brings a challenge in learning spatial features of the output sequence.
On the other hand, it is difficult to learn the differential transition patterns implicitly (without prompting) for the long-horizon prediction conditions. 
Therefore, the proposed differential prompted mechanism significantly contributes to the overall performance of the proposed FlightBERT++. 
In experiment C3, the model suffers from more considerable error accumulation than the original FlightBERT++ with the prediction horizon increasing. 
The weighted sum operation of the ASA module is believed to be more effective in capturing informative flight dynamics and temporal correlations from the observation trajectory sequence.
As can be seen from the results of experiment C4, the model implemented by linear-based TPE is just slightly inferior to the original FlightBERT++ (Conv1D-based TPE), which also exhibits the robustness of the GC representations.

\subsection{Insights} \label{sec5.4}
\subsubsection{Representation ability analysis} \label{sec5.4.1}
\begin{figure*}[htbp] 
        \centering
        \subfloat[FlightBERT++]{\includegraphics[width=2.3 in]{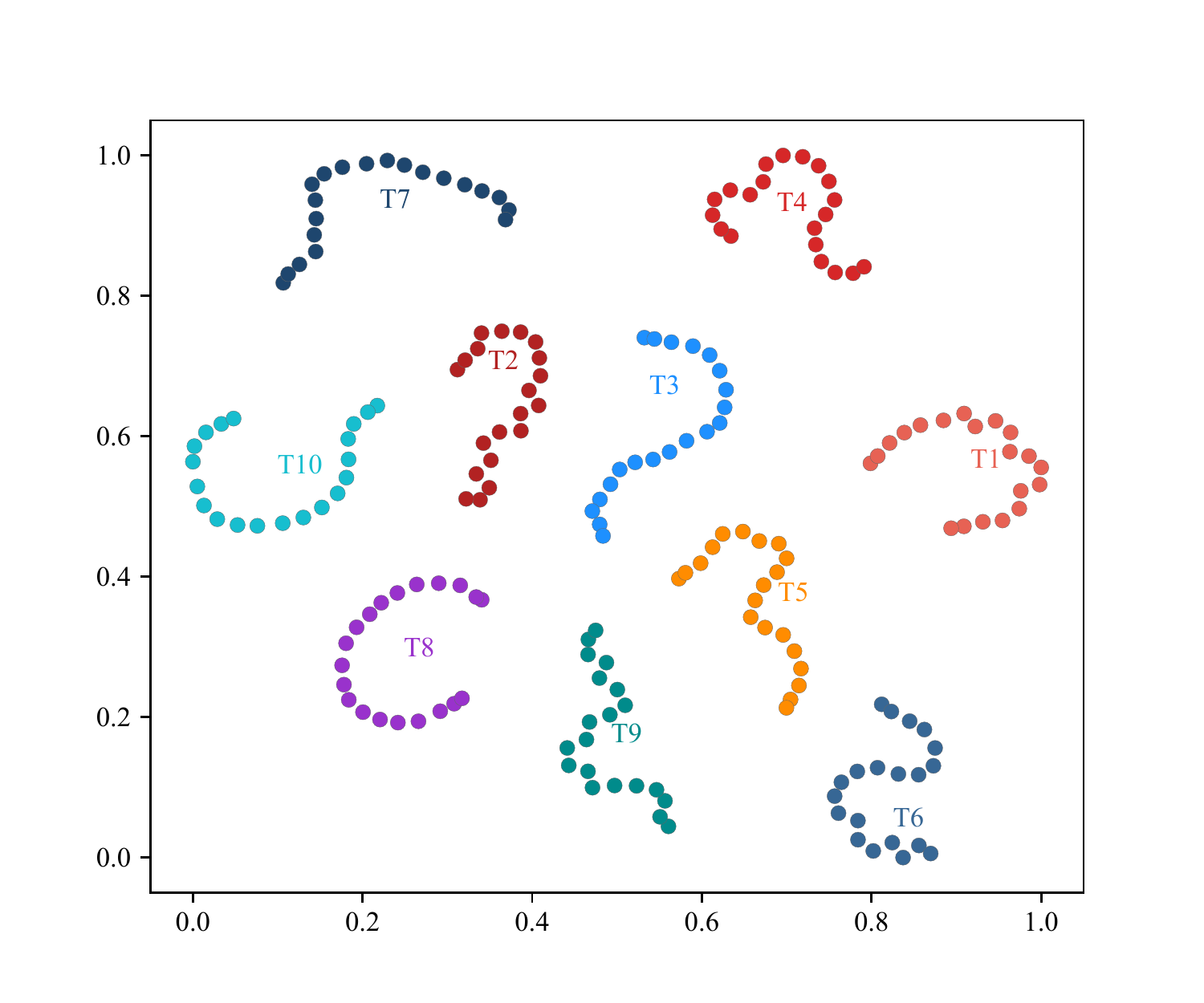} \label{fig_tsne_fb}} %\hspace{0.5 in}
        \subfloat[LSTM+Attention]{\includegraphics[width=2.3 in]{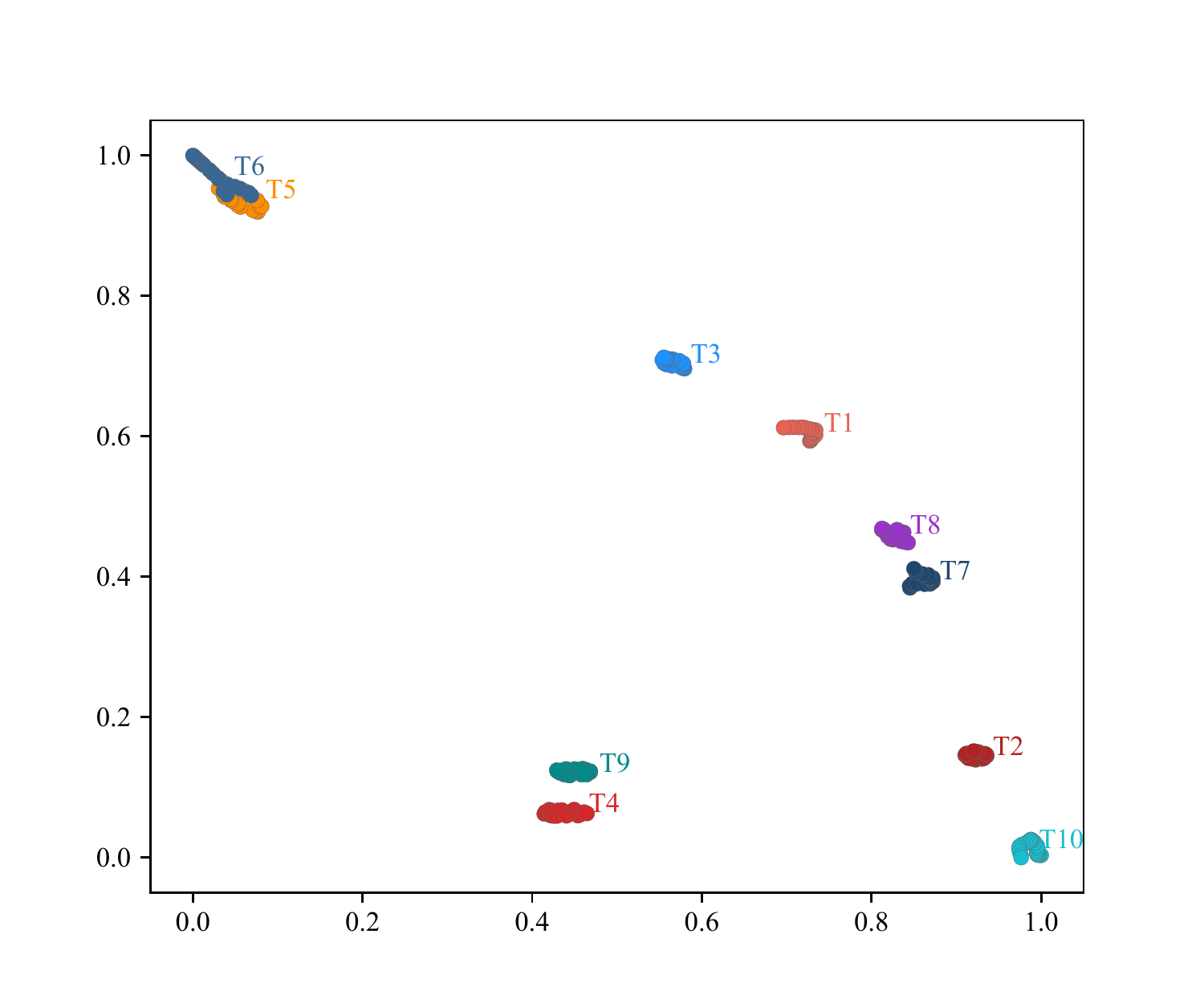} \label{fig_tsne_lstm}} 
        \subfloat[Transformer-Seq2Seq]{\includegraphics[width=2.3 in]{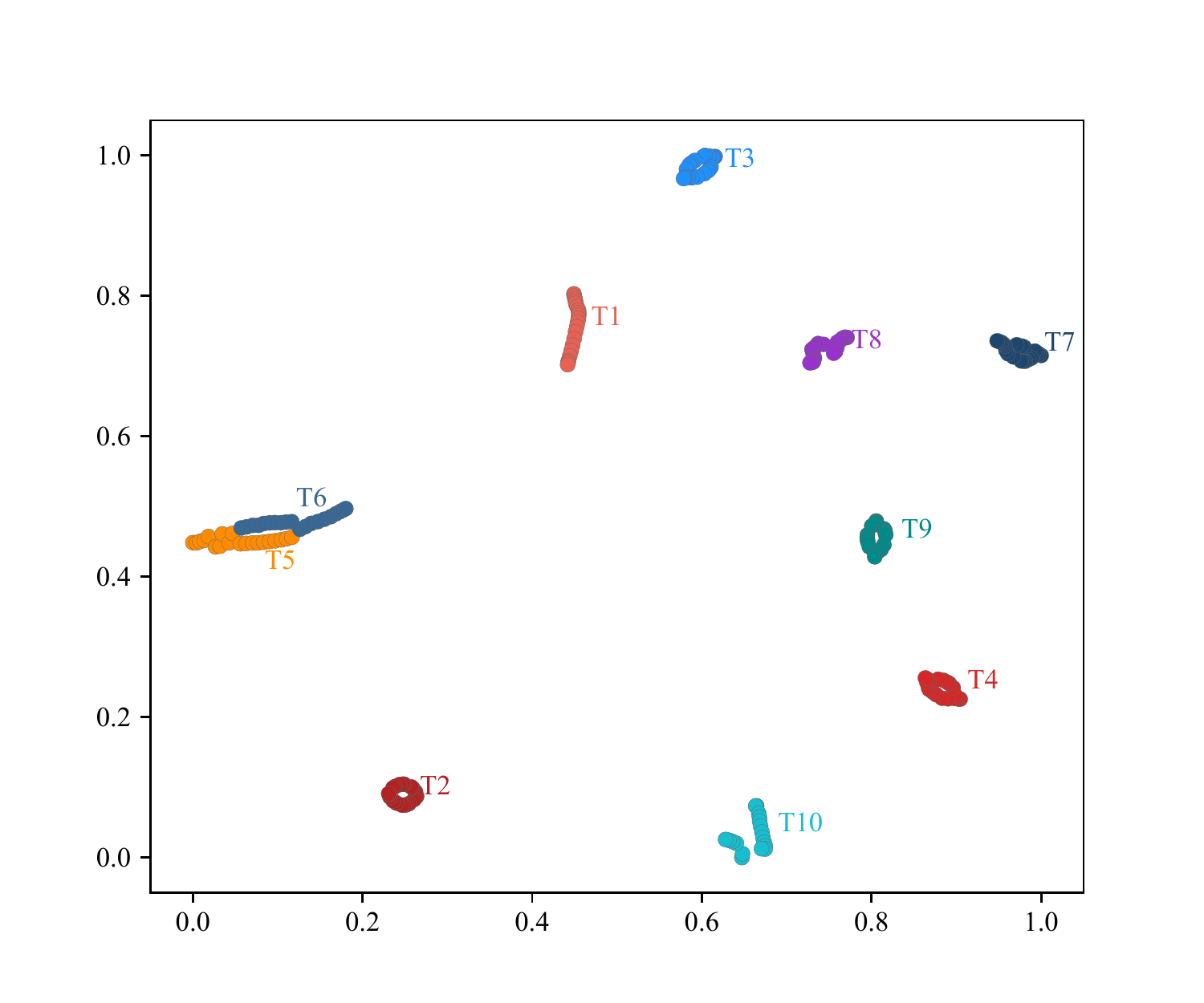} \label{fig_tsne_trans}} 
        \caption{Visualization of the trajectory embeddings in 2D space via the TSNE tool.} \label{fig_tsne}
\end{figure*} 

To better understand the representation ability of the proposed FlightBERT++, a t-Distributed Stochastic Neighbor Embedding (TSNE) tool is employed to visualize the trajectory embeddings in a two-dimensional (2D) space. 
Specifically, we randomly selected 10 flight trajectories (T1 to T10 marked in Figure \ref{fig_tsne}) in the test set. 
In succession, a total of 20 trajectory segments (containing 9 trajectory points for each segment) are extracted from each flight trajectory with a shift length of 1. 
The trajectory segments are further fed into the FTP models to extract the trajectory-level embeddings (e.g., $\mathbf{Traj}_{enc}$) and map these embeddings into 2D space by the TSNE tool. 

The visualization of the trajectory embeddings extracted by the FlightBERT++, LSTM+Attention and Transformer-Seq2Seq models are presented in Figure \ref{fig_tsne_fb}, \ref{fig_tsne_lstm} and \ref{fig_tsne_trans}, respectively. 
In Figure \ref{fig_tsne}, each colored point denotes a certain trajectory embedding, while different trajectories are depicted by individual colors.  
It is clear that the FlightBERT++ can capture discriminative features for different flight trajectories. 
Impressively, the proposed FlightBERT++ learns strong discriminative temporal features among different trajectory segments (even with minor changes of 1 shift length), as well as distinguishes the trajectory embeddings clearly in embedding space among different trajectories.

In contrast, the LSTM+Attention and the Transformer-Seq2Seq models only distinguish the embeddings of the different trajectories but hard to provide the required temporal discriminability among similar trajectory segments in the embedding space. 
In summary, the visualization clearly illustrates that the proposed FlightBERT++ has a stronger representation ability of trajectory sequence, which also provides interpretability for the superior performance of the proposed approach.

\subsubsection{Error analysis (FlightBERT++ v.s. FlightBERT)} \label{sec5.4.2}
\begin{figure}[htbp] 
        \centering 
        \includegraphics[width=3.3 in]{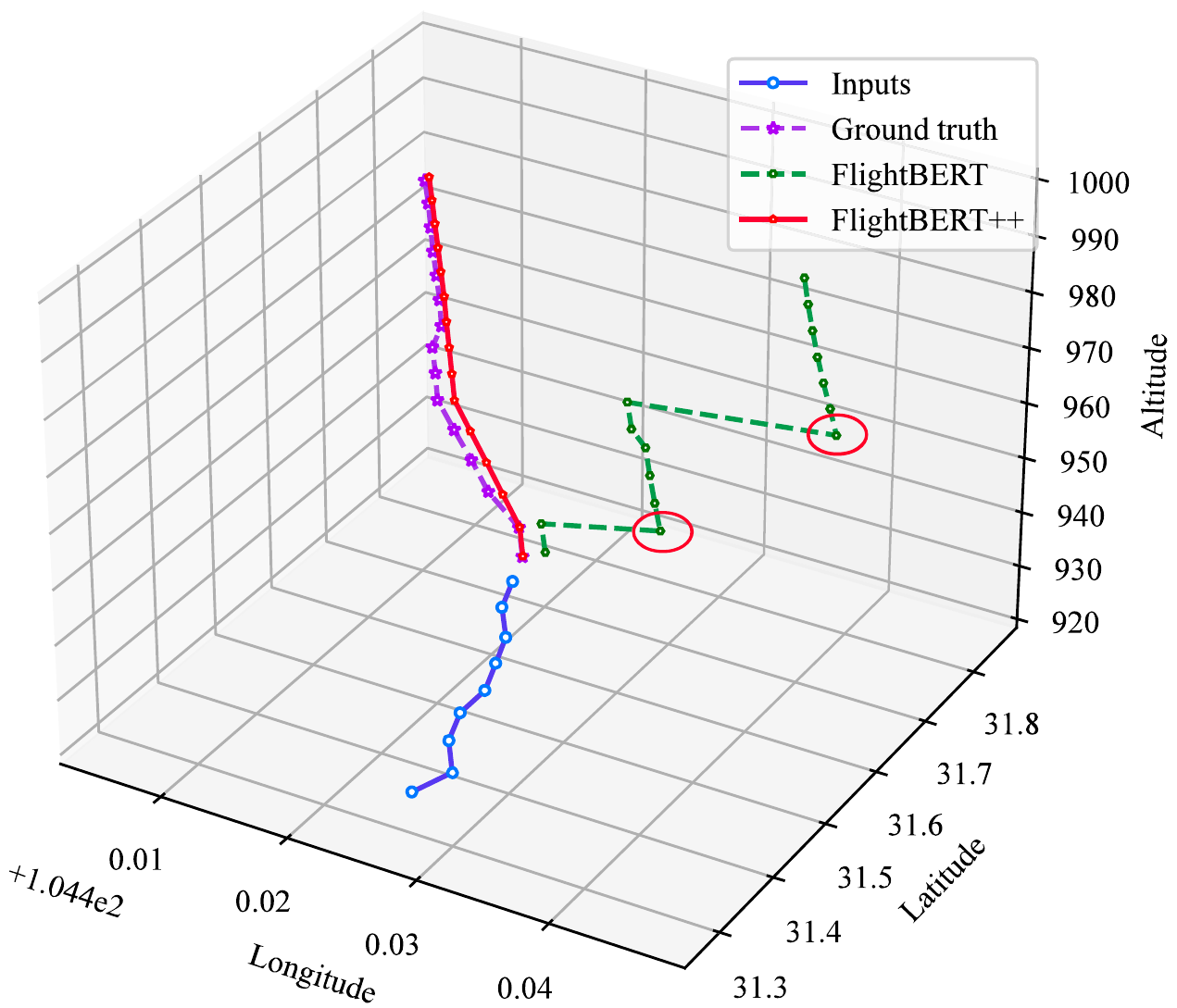}
        \caption{An example of the outliers in FlightBERT caused by high-bit misclassification. The outliers are marked using the red circle.} \label{fig_outlier}
\end{figure}

\begin{figure*}[htbp] 
        \centering 
        \includegraphics[width=7 in]{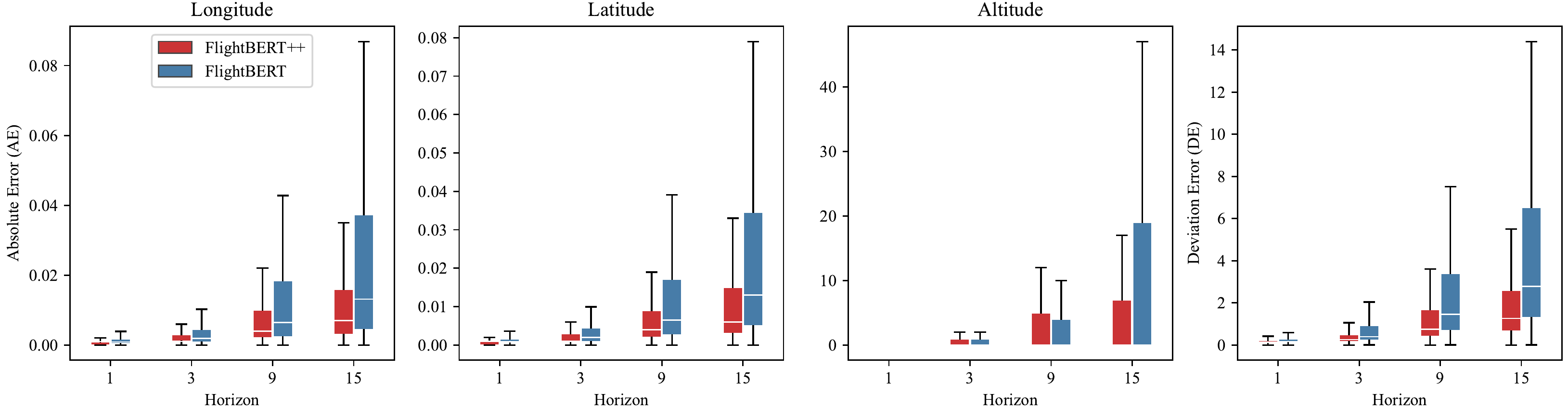} 
        \caption{The boxplots of the absolute error and deviation error for FlightBERT and FlightBERT++.} \label{fig_box_ae}
\end{figure*}

To intuitively illustrate the outliers caused by high-bit misclassification in the BE representation, a case study of the multi-horizon prediction by FlightBERT and FlightBERT++ is visualized in Figure \ref{fig_outlier}. 
Obviously, the high-bit misclassifications occurred in the longitudinal dimension during the $3^{rd}$ and $10^{th}$ prediction horizons, leading to huge prediction errors in the following horizons. 
In contrast, the FlightBERT++ performs the prediction more precisely by incorporating the proposed non-autoregressive multi-horizon forecasting mechanism, differential prediction, and GC representations.   
To further investigate the predicted outliers by FlightBERT and FlightBERT++, the absolute errors of LLA attributes and deviation errors are presented by a set of boxplots in Figure \ref{fig_box_ae}. 
The prediction errors of $1^{st}, 3^{rd}, 9^{th}, 15^{th}$ output by the FlightBERT and FlightBERT++ are illustrated in the boxplots, respectively. 
In order to present the box and whisker of each prediction horizon more clearly, the outliers are removed from the Figures due to the larger number of samples in the test set.

\begin{figure}[htbp] 
        \centering 
        \includegraphics[width=3.45 in]{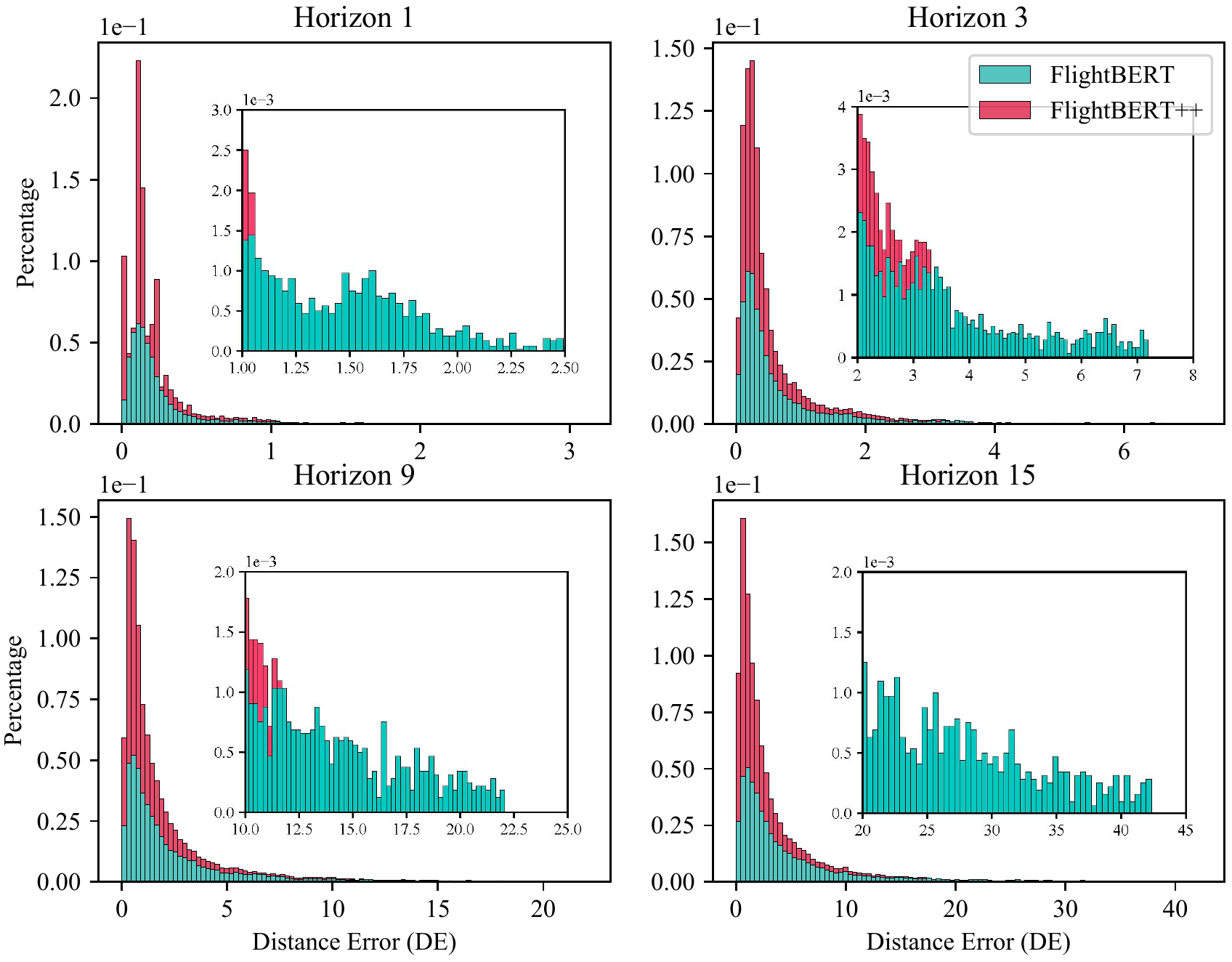}
        \caption{The comparison of the deviation error between FlightBERT and FlightBERT++ via histogram.} \label{fig_his_de}
\end{figure}

As can be seen from the boxplots, compared to the FlightBERT, FlightBERT++ achieves significant performance improvements, especially in longer prediction horizons. 
By analyzing the experimental results, it is found that the FlightBERT model outputs a considerable number of outliers, leading to a higher RMSE. 
To validate this, the distribution of the deviation errors between the FlightBERT and FlightBERT++ is visualized by a histogram in Figure \ref{fig_his_de}. 
It is evident that the FlightBERT framework still outputs larger deviation error values in a certain number of samples. 
In contrast, thanks to the design of the GC representation and differential prediction paradigm, the predictions of the FlightBERT++ show a significant reduction in outliers, which further supports our motivation to address the high-bit misclassifications of the BE representation. 

\subsubsection{Evaluation in different flight phases} \label{sec5.4.3}
To explore the robustness of the proposed FlightBERT++ in different flight phases, the trajectory segments in the test set are further categorized into three subsets according to the flight trends, including climbing, descending/approaching, and en-route. 
In general, the flight transition patterns in the en-route phase are close to the linear, while the climbing and descending/approaching phases are considered as maneuvering operations. 

The experimental results are illustrated in Table \ref{tab_df}. 
As expected, the FlightBERT++ achieves superior performance in the en-route phase, particularly in the attribute of altitude, which benefits from the linear motion patterns in this phase. 
It can be primarily attributed to two key factors: i) The transition patterns in the en-route phase are relatively linear and easy to capture, resulting in accurate predictions.  
ii) The training set comprises a large proportion of en-route samples, as most flights spend a considerable duration in the en-route phase during the entire flight procedure.
It is also observed that the performance of the model suffers from a slight reduction in the climbing and descending phase due to the maneuvering operations.  
In summary, the proposed FlightBERT++ demonstrates the ability to effectively capture different flight dynamics and perform the FTP task across various flight phases, 
allowing it a promising and versatile approach in ATC- and FTP-related applications.

\begin{table*}[htbp]
        \centering
        \setlength\tabcolsep{7.5 pt} 
        \caption{The experimental results evaluated in different flight phases of proposed FlightBERT++.} \label{tab_df}
        \begin{tabular}{c|c|ccc|ccc|ccc|c}
        \hline \hline
        \multirow{2}{*}{\textbf{Flight Phase}} & \multirow{2}{*}{\textbf{Horizon}} & \multicolumn{3}{c|}{\textbf{MAE}} & \multicolumn{3}{c|}{\textbf{MAPE (\%)}} & \multicolumn{3}{c|}{\textbf{RMSE}} & \multirow{2}{*}{\textbf{MDE}} \\ \cline{3-11}
                                               &                                   & Lon      & Lat     & Alt      & Lon        & Lat           & Alt    & Lon       & Lat    & Alt     &                               \\ \hline
        \multirow{4}{*}{Climbing}              & 1                                 & 0.0018   & 0.0020  & 4.68	   & 0.0017	& 0.0072        &1.10	 & 0.0031    & 0.0036 & 6.59    & 0.34                          \\
                                               & 3                                 & 0.0034   & 0.0042  & 8.22	   & 0.0032	& 0.0155	&1.76	 & 0.0060    & 0.0081 & 11.76   & 0.66                          \\
                                               & 9                                 & 0.0095   & 0.0118  & 16.43    & 0.0090     & 0.0435        &2.96	 & 0.0178    & 0.0224 & 23.67   & 1.81                         \\
                                               & 15                                & 0.0152   & 0.0178  & 23.89    & 0.0143     & 0.0656        &3.85	 & 0.0291    & 0.0319 & 34.42   & 2.80                          \\ \hline
        \multirow{4}{*}{En-route}              & 1                                 & 0.0011   & 0.0011  & 0.03	   & 0.0011	& 0.0043	&0.0033	 & 0.0049    & 0.0124 & 0.33    & 0.20                          \\
                                               & 3                                 & 0.0020   & 0.0019  & 0.07	   & 0.0018	& 0.0074	&0.0075	 & 0.0071    & 0.0138 & 0.83    & 0.33                          \\
                                               & 9                                 & 0.0044   & 0.0043  & 0.20	   & 0.0041	& 0.0165	&0.0231	 & 0.0131    & 0.0197 & 2.90    & 0.73                         \\
                                               & 15                                & 0.0068   & 0.0066  & 0.32	   & 0.0063	& 0.0251	&0.0341	 & 0.0185    & 0.0258 & 5.84    & 1.11                          \\ \hline
        \multirow{4}{*}{\makecell{Descending\\Approaching}}            & 1         & 0.0012   & 0.0014  & 2.08	   & 0.0011	& 0.0049	&0.42	 & 0.0027    & 0.0024 & 3.51    & 0.23                         \\
                                               & 3                                 & 0.0026   & 0.0028  & 4.32	   & 0.0025	& 0.0098	&0.90	 & 0.0074    & 0.0055 & 7.73    & 0.46                          \\
                                               & 9                                 & 0.0089   & 0.0083  & 12.14    & 0.0085     & 0.0290        &2.67	 & 0.0248    & 0.0189 & 20.34   & 1.43                         \\
                                               & 15                                & 0.0149   & 0.0135  & 18.23    & 0.0142     & 0.0470        &4.43	 & 0.0387    & 0.0271 & 27.94   & 2.33                          \\ \hline \hline
        \end{tabular}
\end{table*}

\section{Conclusion and Future Work} \label{sec6}
In this paper, we present a novel FTP framework, called FlightBERT++, to perform the multi-horizon trajectory prediction in a non-autoregressive manner. 
The proposed framework not only inherits the superior representation ability of the BE in the FlightBERT framework but also develops an innovative multi-horizon neural network architecture for FTP tasks. 
The FlightBERT++ framework is implemented by a generalized encoder-decoder architecture, in which an additional HACG is designed to generate the multi-horizon contexts by considering the prior horizon information. 
Benefiting from the proposed GC representation and differential prediction paradigm, the FlightBERT++ can mitigate the high-bit misclassification of the BE representation and achieve a significant reduction in outliers than that of the FlightBERT.  
Furthermore, FlightBERT++ exhibits the outstanding capability of trajectory representation through the visualizations of the trajectory embeddings, which provides interpretability for its superior performance. 
In addition, the proposed differential prompt mechanism is also confirmed to contribute to the performance improvements of the FTP task. 
In summary, the proposed framework achieves significant performance improvements and outperforms the comparison baselines across most evaluation metrics, especially in multi-horizon prediction scenarios. 

In the future, we will further focus on technical improvements of the GC representation and the FTP tasks. 
In addition, the practical applications of the FTP in the ATC domain are also interesting research topics, such as conflict detection, and traffic flow prediction.

% {\appendix[Proof of the Zonklar Equations]
% Use $\backslash${\tt{appendix}} if you have a single appendix:
% Do not use $\backslash${\tt{section}} anymore after $\backslash${\tt{appendix}}, only $\backslash${\tt{section*}}.
% If you have multiple appendixes use $\backslash${\tt{appendices}} then use $\backslash${\tt{section}} to start each appendix.
% You must declare a $\backslash${\tt{section}} before using any $\backslash${\tt{subsection}} or using $\backslash${\tt{label}} ($\backslash${\tt{appendices}} by itself
%  starts a section numbered zero.)}


% \bibliography{IEEEabrv,mybibtex}

% You can manually copy in the resultant .bbl file and set second argument of $\backslash${\tt{begin}} to the number of references
%  (used to reserve space for the reference number labels box).

\bibliographystyle{IEEEtran}
\bibliography{IEEEabrv,mybibtex}

\end{document}